%% file: main.tex
\documentclass{article}

\usepackage[preprint,nonatbib]{neurips_2022}

\usepackage[utf8]{inputenc} 
\usepackage[T1]{fontenc}    
\usepackage{hyperref}       
\usepackage{url}            
\usepackage{booktabs}       
\usepackage{amsfonts}       
\usepackage{nicefrac}       
\usepackage{microtype}      
\usepackage{xcolor}         

\usepackage{tabularx}
\usepackage{graphicx}
\usepackage{multirow}
\usepackage{makecell}
\usepackage{caption}
\usepackage{subcaption}
\usepackage{soul} 
\usepackage{amsmath}
\usepackage{listings}
\usepackage{colortbl}  
\usepackage[normalem]{ulem}
\useunder{\uline}{\ul}{}
\usepackage{CJKutf8} 
\usepackage[misc]{ifsym}

\definecolor{codegreen}{rgb}{0,0.6,0}
\definecolor{codegray}{rgb}{0.5,0.5,0.5}
\definecolor{codepurple}{rgb}{0.58,0,0.82}
\definecolor{backcolour}{rgb}{0.95,0.95,0.92}

\lstdefinestyle{mystyle}{
    backgroundcolor=\color{backcolour},   
    commentstyle=\color{codegreen},
    keywordstyle=\color{magenta},
    numberstyle=\tiny\color{codegray},
    stringstyle=\color{codepurple},
    basicstyle=\ttfamily\footnotesize,
    breakatwhitespace=false,         
    breaklines=true,                 
    captionpos=b,                    
    keepspaces=true,                 
    numbers=left,                    
    numbersep=5pt,                  
    showspaces=false,                
    showstringspaces=false,
    showtabs=false,                  
    tabsize=2
}

\title{How Close is ChatGPT to Human Experts? \\Comparison Corpus, Evaluation, and Detection}

\author{%
Biyang Guo$^{1\dag}$\thanks{Equal Contribution.}~~, Xin Zhang$^{2*}$, Ziyuan Wang$^{1*}$, Minqi Jiang$^{1*}$, Jinran Nie$^{3*}$ \\
\textbf{Yuxuan Ding$^{4}$, Jianwei Yue$^{5}$, Yupeng Wu$^{6}$} \\
$^1$AI Lab, School of Information Management and Engineering \\Shanghai University of Finance and Economics\\
$^2$Institute of Computing and Intelligence, Harbin Institute of Technology (Shenzhen) \\
$^3$School of Information Science, Beijing Language and Culture University\\
$^4$School of Electronic Engineering, Xidian University \\
$^5$School of Computing, Queen's University, $^6$Wind Information Co., Ltd \\
}

\definecolor{antiquefuchsia}{rgb}{0.57, 0.36, 0.51}

\begin{document}

\maketitle

\begingroup\def\thefootnote{$^\dag$}\footnotetext{Project Lead. Corresponding to \texttt{guo\_biyang@163.com}}\endgroup
\begingroup\def\thefootnote{$^+$}\footnotetext{Each author has made unique contributions to the project.}\endgroup
\begin{abstract}
  The introduction of ChatGPT\footnote{Launched by OpenAI in November 2022. \url{https://chat.openai.com/chat}} has garnered widespread attention in both academic and industrial communities. ChatGPT is able to respond effectively to a wide range of human questions, providing fluent and comprehensive answers that significantly surpass previous public chatbots in terms of security and usefulness. On one hand, people are curious about how ChatGPT is able to achieve such strength and how far it is from human experts. On the other hand, people are starting to worry about the potential negative impacts that large language models (LLMs) like ChatGPT could have on society, such as fake news, plagiarism, and social security issues. In this work, we collected tens of thousands of comparison responses from both human experts and ChatGPT, with questions ranging from open-domain, financial, medical, legal, and psychological areas. We call the collected dataset the \textbf{H}uman \textbf{C}hatGPT \textbf{C}omparison \textbf{C}orpus (\textbf{HC3}). Based on the HC3 dataset, we study the characteristics of ChatGPT's responses, the differences and gaps from human experts, and future directions for LLMs. We conducted comprehensive human evaluations and linguistic analyses of ChatGPT-generated content compared with that of humans, where many interesting results are revealed. After that, we conduct extensive experiments on how to effectively detect whether a certain text is generated by ChatGPT or humans. We build three different detection systems, explore several key factors that influence their effectiveness, and evaluate them in different scenarios.  The dataset, code, and models are all publicly available at \url{https://github.com/Hello-SimpleAI/chatgpt-comparison-detection}.
\end{abstract}

\section{Introduction}

Since its dazzling debut in November 2022, OpenAI's ChatGPT has gained huge attention and wide discussion in the natural language processing (NLP) community and many other fields. According to OpenAI, ChatGPT is fine-tuned from the GPT-3.5 series with Reinforcement Learning from Human Feedback (RLHF; \cite{PaulFChristiano2017-RLHF-1,NisanStiennon2020LearningTS-RLHF-2}), using nearly the same methods as InstructGPT \cite{ouyang2022training-InstructGPT}, but with slight differences in the data collection setup. 
The vast amount of knowledge in GPT-3.5 and the meticulous fine-tuning based on human feedback enable ChatGPT to excel at many challenging NLP tasks, such as translating natural language to code \cite{chen2021evaluating-CodeX}, completing the extremely masked text \cite{guo2022genius} or generating stories given user-defined elements and styles \cite{yao2019plan-and-write}, let alone typical NLP tasks like text classification, entity extraction, translation, etc. Furthermore, the carefully collected human-written demonstrations also make ChatGPT able to admit its mistakes, challenge incorrect premises and reject even inappropriate requests, as claimed by OpenAI\footnote{\url{https://openai.com/blog/chatgpt/}}.


The surprisingly strong capabilities of ChatGPT have raised many interests, as well as concerns:

On the one hand, \textbf{people are curious about how close is ChatGPT to human experts}. Different from previous LLMs like GPT-3 \cite{brown2020language-gpt3}, which usually fails to properly respond to human queries, InstructGPT \cite{ouyang2022training-InstructGPT} and the stronger ChatGPT have improved greatly in interactions with humans. Therefore, ChatGPT has great potential to become a daily assistant for general or professional consulting purposes \cite{jeblick2022chatgpt-medical-report,king2022-med-future}. From the linguistic or NLP perspectives, we are also interested in where are the remaining gaps between ChatGPT and humans and what are their implicit linguistic differences \cite{goldstein2022shared-lm-human,hu2022fine-ling-human-vs-lm}.

On the other hand, \textbf{people are worried about the potential risks brought by LLMs like ChatGPT}. With the free preview demo of ChatGPT going virus, a large amount of ChatGPT-generated content crowded into all kinds of UGC (User-Generated Content) platforms, threatening the quality and reliability of the platforms. For example, Stack Overflow, the famous programming question-answering website, has temporarily banned ChatGPT-generated content\footnote{
\url{https://meta.stackoverflow.com/questions/421831/temporary-policy-chatgpt-is-banned}
}, because it believes \textit{"the average rate of getting correct answers from ChatGPT is too low, the posting of answers created by ChatGPT is substantially harmful to the site and to users who are asking and looking for correct answers"}. Many other applications and activities are facing similar issues, such as online exams \cite{susnjak2022-online-exam} and medical analysis \cite{jeblick2022chatgpt-medical-report}. Our empirical evaluation of ChatGPT on legal, medical, and financial questions also reveals that potentially harmful or fake information can be generated. 

Considering the opaqueness of ChatGPT and the potential social risks associated with model misuse, we make the following contributions to both the academy and society:
\begin{itemize}
    \item[1.] To facilitate LLM-related research, especially the study on the comparison between humans and LLMs, we collect nearly 40K questions and their corresponding answers from human experts and ChatGPT, covering a wide range of domains (open-domain, computer science, finance, medicine, law, and psychology), named as the \textbf{Human ChatGPT Comparison Corpus} (\textbf{HC3}) dataset. The HC3 dataset is a valuable resource to analyze the linguistic and stylist characteristics of both humans and ChatGPT, which helps to investigate the future improvement directions for LLMs;
    \item[2.] We conduct comprehensive \textbf{human evaluations} as well as \textbf{linguistic analysis} on human/ChatGPT-generated answers, discovering many interesting patterns exhibited by humans and ChatGPT. These findings can help to distinguish whether certain content is generated by LLMs, and also provide insights about where language models should be heading in the future;
    \item[3.] Based on the HC3 dataset and the analysis, we develop several \textbf{ChatGPT detecting models}, targeting different detection scenarios. These detectors show decent performance in our held-out test sets. We also conclude several key factors that are essential to the detector's effectiveness.
    \item[4.] We \textbf{open-source} all the collected comparison corpus, evaluations, and detection models, to facilitate future academic research and online platform regulations on AI-generated content.
\end{itemize}


\section{Human ChatGPT Comparison Corpus (HC3)}
ChatGPT is based on the GPT-3.5 series, which is pre-trained on the super-large corpus, consisting of web-crawled text, books, and codes, 
making it able to respond to all kinds of questions. Therefore, we are curious how will a human (especially an expert) and ChatGPT respond to the same question respectively. Inspired by \cite{askell2021general}, we also want to evaluate whether ChatGPT can keep honest (not fabricate information or mislead the user), harmless (shouldn't generate harmful or offensive content), and how \textit{helpful} (provide concrete and correct solutions to the user's question) it is compared to human experts.

Taking these into account, we decided to collect a comparison corpus that consists of both human and ChatGPT answers to the same questions. We believe such a comparison corpus can be a valuable and interesting source to study the nature of the language of both humans and language models.

\input{tables/dataset_meta-biyang.tex}

\subsection{Human Answers Collection}
Inviting human experts to manually write questions and answers is tedious and unaffordable for us to collect a large amount of data, therefore we construct the comparison dataset mainly from two sources:\\
$\bullet$ Publicly available question-answering datasets, where answers are given by experts in specific domains or the high-voted answers by web users;\\
$\bullet$ Wiki text. We construct question-answer pairs using the concepts and explanations from wiki sources like Wikipedia\footnote{\url{https://www.wikipedia.org/}} and BaiduBaike\footnote{\url{https://baike.baidu.com/}}.

The split-data source mapping is shown in Table \ref{dataset_meta-biyang}, and please refer to Appendix \ref{app:splits} for further detailed information.

\subsection{ChatGPT Answers Collection}
Based on the collected human question-answering datasets, we use ChatGPT to generate answers to these questions. Since the ChatGPT is currently only available through its preview website, we manually input the questions into the input box, and get the answers, with the aid of some automation testing tools. Answers by ChatGPT can be influenced by the chatting history, so we refresh the thread for each question.

To make the answer more aligned with human answers, we add additional instructions to ChatGPT for specific datasets. For example, the human answers from the \texttt{reddit-eli5} dataset split are under the context of "Explain like I'm five", therefore we use this context to instruct ChatGPT by adding "Explain like I'm five" at the end of the original question. More detail can be found in the Appendix.

ChatGPT can generate different answers given the same question in different threads, which is perhaps due to the random sampling in the decoding process. However, we found the differences can be very small, thereby we only collect one answer for most questions.

\subsection{Human ChatGPT Comparison Corpus (HC3)}
For each question, there can be more than one human/ChatGPT answer, therefore we organize the comparison data using the following format:
\begin{lstlisting}
{
    "question": "Q1", 
    "human_answers": ["A1", "A2"], 
    "chatgpt_answers": ["B1"]
}
\end{lstlisting}

Overall, we collected $24,322$ questions, $58,546$ human answers and $26,903$ ChatGPT answers for the English version, and $12,853$ questions, $22,259$ human answers and $17,522$ ChatGPT answers for the Chinese version.
The meta-information of each dataset split is illustrated in Table \ref{dataset_meta-biyang}.

\section{Human Evaluation \& Summarization}\label{sec:human-eval-sum}

In this section, we invite many volunteer testers and conduct extensive human evaluations from different aspects. After the human evaluation, we make our collected comparison corpus available to the volunteers and ask them to manually conclude some characteristics. We then summarize the feedback from the volunteers combined with our observations.

\subsection{Human Evaluation}\label{sec:human}
The human evaluation is divided into the \textbf{Turing test} and the \textbf{Helpfulness Test}. The Turing Test \cite{turing2009-turing-test} is a test of a machine's ability to exhibit intelligent behavior that is indistinguishable from a human. We invite 17 volunteers, divided into two groups: 8 experts (who are frequent users of ChatGPT) and 9 amateurs (who have never heard of ChatGPT). This is because people who are familiar with ChatGPT may have memorized some patterns exhibited by ChatGPT, helping them to easily distinguish the role.

We designed four types of evaluations, using different query formats or testing groups. We introduce the specific evaluation design and results in the following parts:

\textbf{$\mathcal{A.}$ Expert Turing Test, Paired Text (\texttt{pair-expert})}\\
The \texttt{pair-expert} test is conducted in the \textbf{expert} group. Each tester is required to do a series of tests, each test containing one question and a \textbf{pair} of answers (one from humans and another from ChatGPT). The tester needs to determine which answer is generated by ChatGPT.

\textbf{$\mathcal{B.}$ Expert Turing Test, Single Text (\texttt{single-expert})}\\
The \texttt{single-expert} test is also conducted in the \textbf{expert} group. Each tester is required to do a series of tests, each test containing one question and a \textbf{single} answer randomly given by humans or ChatGPT. The tester needs to determine whether the answer is generated by ChatGPT.


\textbf{$\mathcal{C.}$ Amateur Turing Test, Single Text (\texttt{single-amateur})}\\
The \texttt{single-amateur} test is conducted in the \textbf{amateur} group. Each tester is required to do a series of tests, each test containing one question and a \textbf{single} answer randomly given by humans or ChatGPT. The tester needs to determine whether the answer is generated by ChatGPT.


\textbf{$\mathcal{D.}$ Helpfulness Test (\texttt{helpfulness})}\\
We are also curious about how helpful are the answers from ChatGPT compared with humans' answers to one question. Note that helpfulness is a very subjective metric, which can be influenced by many factors, including emotion, tester personality, personal preference, etc. Therefore, simply providing more accurate information or a more detailed analysis may not always lead to a more helpful answer.

The \texttt{helpfulness} test is conducted in the \textbf{expert} group. Each tester is required to do a series of tests, each containing one question and a \textbf{pair} of answers (one from human and another from ChatGPT). Each tester is asked to pretend that the question is proposed by him/herself, and needs to determine which answer is more helpful to him/her.

\paragraph{Settings.} We sample around 30 \texttt{<question, human\_answer, chatgpt\_answer>} triplets from each split (i.e., \textit{reddit\_eli5}, \textit{wikipedia}, \textit{medical}, etc.) as the samples for the human evaluation. We allocate 2-5 testers for each split and report their average results. For all Turing tests, we report \textit{the proportion that ChatGPT-generated answer is correctly detected} by testers. For the helpfulness test, we report \textit{the proportion that ChatGPT-generated answer is considered to be more helpful}.

\input{tables/human_evaluation.tex}

\paragraph{Results.} Several conclusions can be drawn from the results shown in Table \ref{human_eval_all}. Comparing the results of \texttt{pair-expert} and \texttt{single-expert}, we can find that \textbf{it is easier to distinguish ChatGPT-generated content when providing a comparison pair} than only providing a single answer. Comparing the results of \texttt{single-expert} and \texttt{single-amateur}, we can find that \textbf{the accuracy of experts is much higher than that of amateurs}. The \texttt{helpfulness} test gives the proportion of questions that volunteers think the ChatGPT answer is more helpful to them. Surprisingly, results show that \textbf{ChatGPT's answers are generally considered to be more helpful than humans' in more than half of questions}, especially for finance and psychology areas. By checking the specific answers in these domains, we find that ChatGPT can usually provide more concrete and specific suggestions. However, ChatGPT performs poorly in terms of helpfulness for the medical domain in both English and Chinese. The ChatGPT often gives lengthy answers to medical consulting in our collected dataset, while human experts may directly give straightforward answers or suggestions, which may partly explain why volunteers consider human answers to be more helpful in the medical domain.

\subsection{Human Summarization}

After the above evaluations, we open our collected HC3 dataset to the volunteers where they can freely browse the comparison answers from humans and ChatGPT. All dataset splits are allocated to different volunteers, and each volunteer is asked to browse at least 100 groups of comparison data. After that, we ask them to summarize the characteristics of both human answers and ChatGPT answers.
Eventually, we received more than 200 feedbacks, and we summarize these findings as follows:

\textbf{Distinctive Patterns of ChatGPT}
\begin{itemize}
    \item[(a)] \textbf{ChatGPT writes in an organized manner, with clear logic}. Without loss of generality, ChatGPT loves to define the core concept in the question. Then it will give out detailed answers step by step and offers a summary at the end, following the deduction and summary structure;
    \item[(b)] \textbf{ChatGPT tends to offer a long and detailed answer.} This is the direct product of the Reinforcement Learning with Human Feedback, i.e. RLHF, and also partly related to the pattern (a) unless you offer a prompt such as "Explain it to me in one sentence";
    \item[(c)] \textbf{ChatGPT shows less bias and harmful information}. ChatGPT is neutral on sensitive topics, barely showing any attitude towards the realm of politics or discriminatory toxic conversations;
    \item[(d)] \textbf{ChatGPT refuses to answer the question out of its knowledge.} For instance, ChatGPT cannot respond to queries that require information after September 2021. Sometimes ChatGPT also refuses to answer what it believes it doesn't know. It is also RLHF's ability to implicitly and automatically determine which information is within the model's knowledge and which is not.
    \item[(e)] \textbf{ChatGPT may fabricate facts.} When answering a question that requires professional knowledge from a particular field, ChatGPT may fabricate facts in order to give an answer, though \cite{ouyang2022training-InstructGPT} mentions that InstructGPT model has already shown improvements in truthfulness over GPT-3. For example, in legal questions, ChatGPT may invent some non-existent legal provisions to answer the question. This phenomenon warns us to be extra careful when using ChatGPT for professional consultations. Additionally, when a user poses a question that has no existing answer, ChatGPT may also fabricate facts in order to provide a response.

\end{itemize}

Many of the conclusions mentioned above like (b),(c),(d) are also discussed in \cite{fu2022gptroadmap} by Fu et al.

\textbf{Major Differences between Human and ChatGPT}
\begin{itemize}
    \item[(a)] \textbf{ChatGPT's responses are generally strictly focused on the given question, whereas humans' are divergent and easily shift to other topics.} In terms of the richness of content, humans are more divergent in different aspects, while ChatGPT prefers focusing on the question itself. Humans can answer the hidden meaning under the question based on their own common sense and knowledge, but the ChatGPT relies on the literal words of the question at hand;
    \item[(b)] \textbf{ChatGPT provides objective  answers, while humans prefer subjective expressions.} Generally, ChatGPT generates safer, more balanced, neutral, and informative texts compared to humans. As a result, ChatGPT is excellent at interpreting terminology and concepts. On the other hand, human answers are more specific and include detailed citations from sources based on legal provisions, books, and papers, especially when providing suggestions for medical, legal, and technical problems, etc.;    
    \item[(c)] \textbf{ChatGPT's answers are typically formal, meanwhile humans' are more colloquial.} Humans tend to be more succinct with full of oral abbreviations and slang such as "LOL", "TL;DR", "GOAT" etc. Humans also love to apply humor, irony, metaphors, and examples, whereas ChatGPT never uses antiphrasis. Additionally, human communication often includes the "Internet meme" as a way to express themselves in a specific and vivid way;
    \item[(d)] \textbf{ChatGPT expresses less emotion in its responses, while human chooses many punctuation and grammar feature in context to convey their feelings.} Human uses multiple exclamation mark('!'), question mark('?'), ellipsis('...') to express their strong emotion, and use various brackets('(', ')', '[', ']') to explain things. By contrast, ChatGPT likes to use conjunctions and adverbs to convey a logical flow of thought, such as "In general", "on the other hand", "Firstly,..., Secondly,..., Finally" and so on.
\end{itemize}

Overall, these summarised features indicate that ChatGPT has improved notably in question-answering tasks for a wide range of domains. Compared with humans, we can imagine ChatGPT as a conservative \textit{team} of experts. As a "team", it may lack individuality but can have a more comprehensive and neutral view towards questions.

\section{Linguistic Analysis}
In this section, we analyze the linguistic features of both humans' and ChatGPT's answers, and try to find some statistical evidence for the characteristics concluded in Section \ref{sec:human-eval-sum}.

\input{tables/vocab_compare-biyang.tex}

\subsection{Vocabulary Features}
In this part, we analyze the vocabulary features of our collected corpus. We are interested in how humans and ChatGPT differ in the choice of words when answering the same set of questions. 

Since the number of human/ChatGPT answers is unbalanced, we randomly sample one answer from humans and one answer from ChatGPT during our statistical process. We calculated the following features: \textbf{average length ($L$)}, which is the average number of words in each question; \textbf{vocab size ($V$)}, the number of unique words used in all answers; we also propose another feature called \textbf{density ($D$)}, which is calculated by $D = 100\times V/(L\times N)$ where $N$ is the number of answers. Density measures how \textit{crowded} different words are used in the text. For example, if we write some articles that add up to 1000 words, but only 100 different words are used, then the density is $100\times 100/1000=10$. The higher the density is, the more different words are used in the same length of text.

In Table \ref{vocab_compare}, we report the vocabulary features for both English and Chinese corpus.
Looking at both features of \textit{average length} and \textit{vocab size}, we can see that: \textbf{compared to ChatGPT, human answers are relatively shorter, but a larger vocabulary is used.} This phenomenon is particularly obvious in the Chinese \texttt{open\_qa} split and the \texttt{medical} splits in both languages, where the average length of ChatGPT is nearly twice longer than that of humans, but the vocab size is significantly smaller.

This phenomenon is also reflected by the \textit{density} factor. The word density of humans is greater than ChatGPT's in \textbf{every split}, which further reveals that \textbf{humans use a more diverse vocabulary in their expressions}.

\subsection{Part-of-Speech \& Dependency Analysis}

In this part, we compare the occurrences of different part-of-speech (POS) tags and the characteristics of the dependency relations.

\subsubsection{Part-of-Speech}
\begin{figure}[t]
	\centering
	\begin{minipage}[t]{0.95\textwidth}
		\centering
		\includegraphics[width=\textwidth]{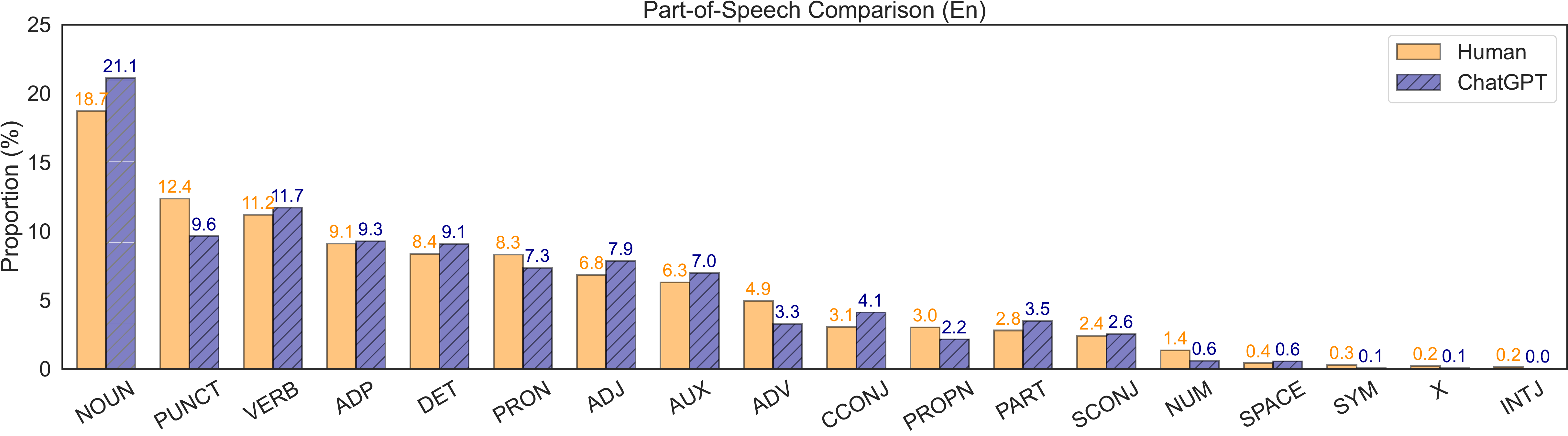}
	\end{minipage}
	\begin{minipage}[t]{0.95\textwidth}
		\centering
		\includegraphics[width=\textwidth]{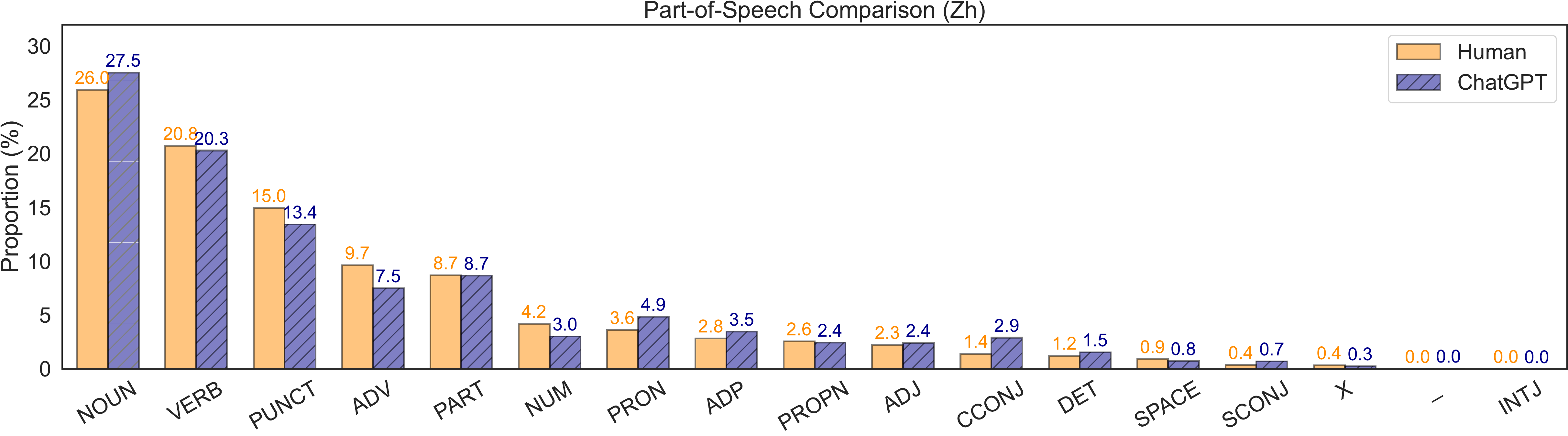}
	\end{minipage}
	\caption{Part-of-Speech distribution comparison between ChatGPT and human answers. Results are sorted by POS proportion of human answers. The upper figure is for the HC3-English dataset and the lower is for the HC3-Chinese dataset.}
 \label{fig:pos-yuxuan}
\end{figure}

Figure \ref{fig:pos-yuxuan} illustrates the comparisons between humans and ChatGPT in terms of POS usage. In HC3-English, ChatGPT uses \textbf{more} \texttt{NOUN}, \texttt{VERB}, \texttt{DET}, \texttt{ADJ}, \texttt{AUX}, \texttt{CCONJ} and \texttt{PART} words, while using less \texttt{ADV} and \texttt{PUNCT} words. 

 A high proportion of nouns (\texttt{NOUN}) often indicates that the text is more argumentative, exhibiting informativeness and objectivity \cite{nagy2012pos-analysis-1}. Accordingly, adposition (\texttt{ADP}) and adjective (\texttt{ADJ}) words also tend to appear more frequently \cite{fang2006-pos-analysis-2}. 
 The frequent co-occurrence of conjunctions (\texttt{CCONJ}) along with nouns, verbs, and adposition words indicates that the structure of the article and the relationships of cause-and-effect, progression, or contrast are clear. The above are also typical characteristics in academic papers or official documents \cite{schleppegrell2004language-pos-analysis-3}. We believe the RLHF training process has a great influence on ChatGPT's writing style, which partly explains the difference in the POS tags distribution.

\subsubsection{Dependency Parsing}
\begin{figure}[t]
	\centering
	\begin{minipage}[t]{\textwidth}
		\centering
		\includegraphics[width=\textwidth]{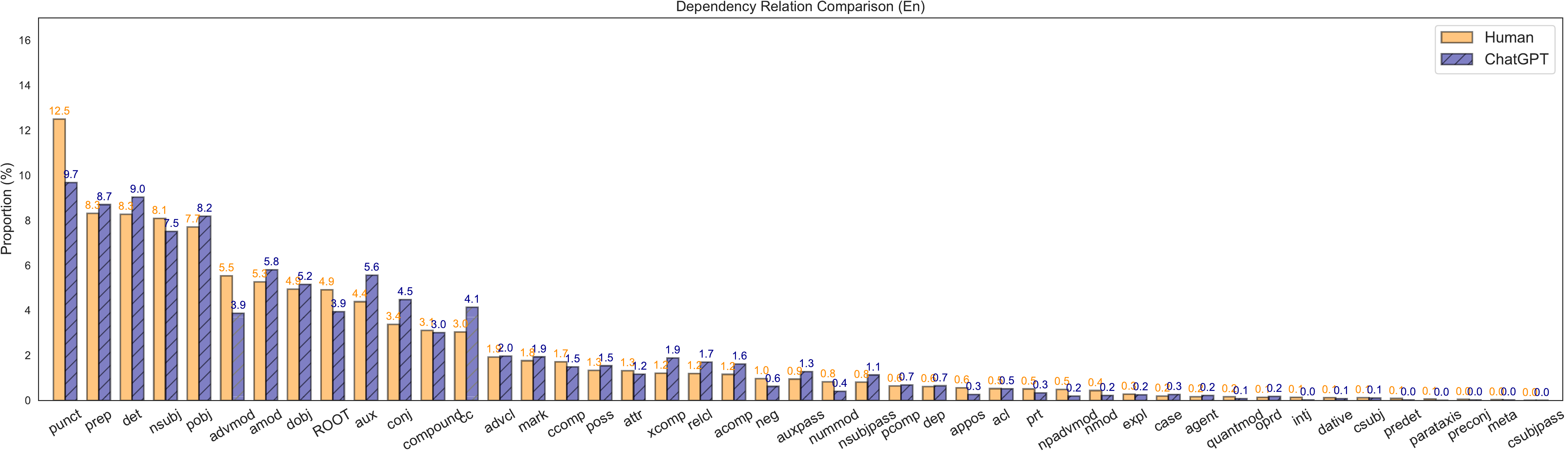}
	\end{minipage}
	\begin{minipage}[t]{\textwidth}
		\centering
		\includegraphics[width=\textwidth]{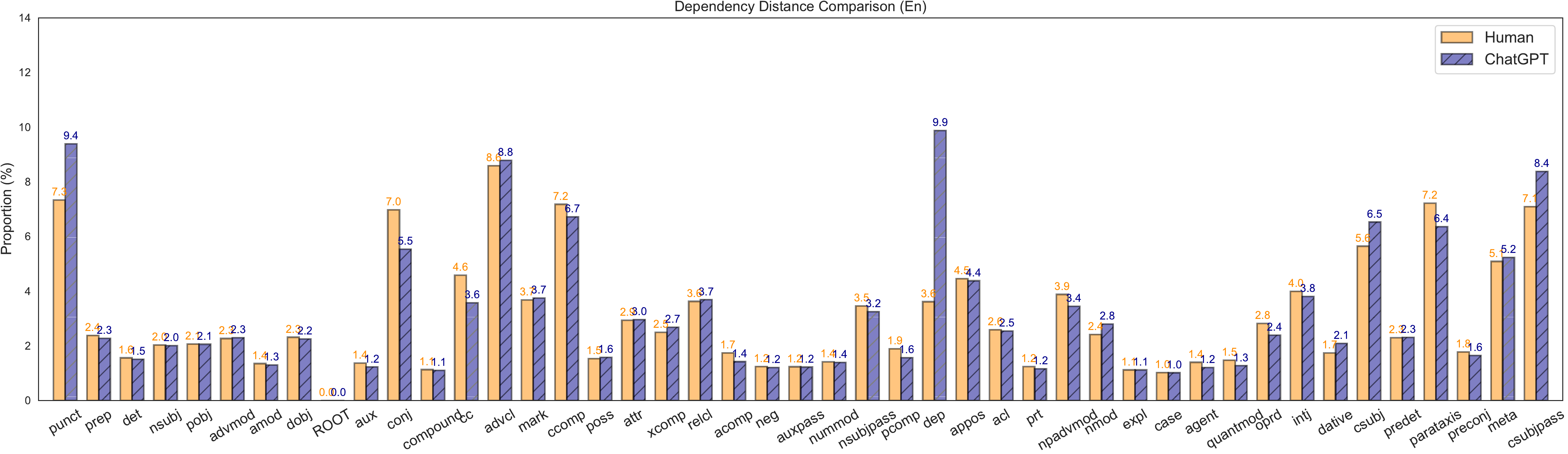}
	\end{minipage}
	\caption{Top-30 dependency relations  (upper) and corresponding dependency distances (lower) comparison between human and ChatGPT answers in HC3-English. Results are sorted by relations proportion of human answers.}
 \label{fig:dep-and-dist-yuxuan}
\end{figure}

Dependency parsing is a technique that analyzes the grammatical structure of a sentence by identifying the dependencies between its words. We parse the answers in the corpus and compare the proportion of different dependency relations and their corresponding dependency distances. Figure \ref{fig:dep-and-dist-yuxuan} shows the comparison between humans and ChatGPT in HC3-English. Due to the limited space, the Chinese version is placed in the Appendix \ref{app: depency}.

The comparison of dependency relations exhibits similar characteristics to that of POS tags, where ChatGPT uses more determination, conjunction, and auxiliary relations. In terms of the dependency distance, ChatGPT has much longer distances for the \texttt{punct} and \texttt{dep} relations, which is perhaps due to the fact that CharGPT tends to use longer sentences. However, ChatGPT has obviously shorter \texttt{conj} relations. According to the analysis of POS tags, ChatGPT usually uses more conjunctions than humans to make the content more logical, this may explain why the \texttt{conj} relations of ChatGPT are relatively shorter than humans.

\subsection{Sentiment Analysis}
Humans are emotional beings, it is natural that our emotions are reflected in our words, to some extent. ChatGPT is learned on large-scale human-generated text, but it is further fine-tuned with human instructions. Therefore we are curious how "emotional" ChatGPT is compared with humans.

We use a multilingual sentiment classification model\footnote{https://huggingface.co/cardiffnlp/twitter-xlm-roberta-base-sentiment} fine-tuned on Twitter corpus \cite{twitter-sentiment-clf} to conduct sentiment analysis for both English and Chinese comparison data. Note that deep learning-based models can be greatly influenced by some indicating words (such as "but" and "sorry" can easily fool the classifier to predict the "negative" label), making the predictions biased \cite{guo2022selective-STA}. Therefore, the sentiment given by the classifier is only a reference to the true sentiment behind the text.

\begin{figure*}[t]
 \centering
  \begin{subfigure}[b]{0.49\textwidth}
         \centering
         \includegraphics[width=\textwidth]{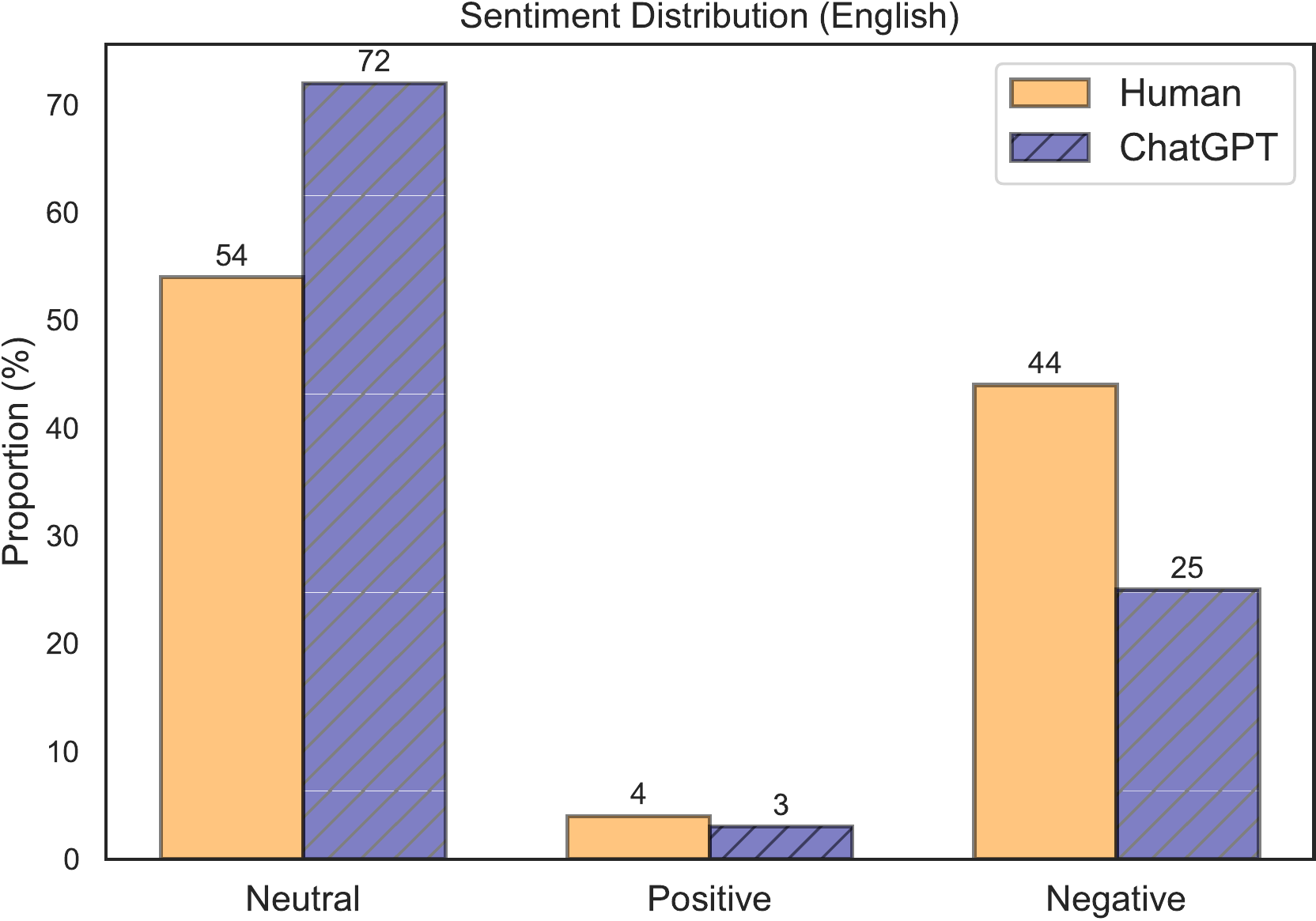}
         \caption{Sentiment distribution of HC3-English}
         \label{fig:en-sentiment}
     \end{subfigure}
\hfill
     \begin{subfigure}[b]{0.49\textwidth}
         \centering
         \includegraphics[width=\textwidth]{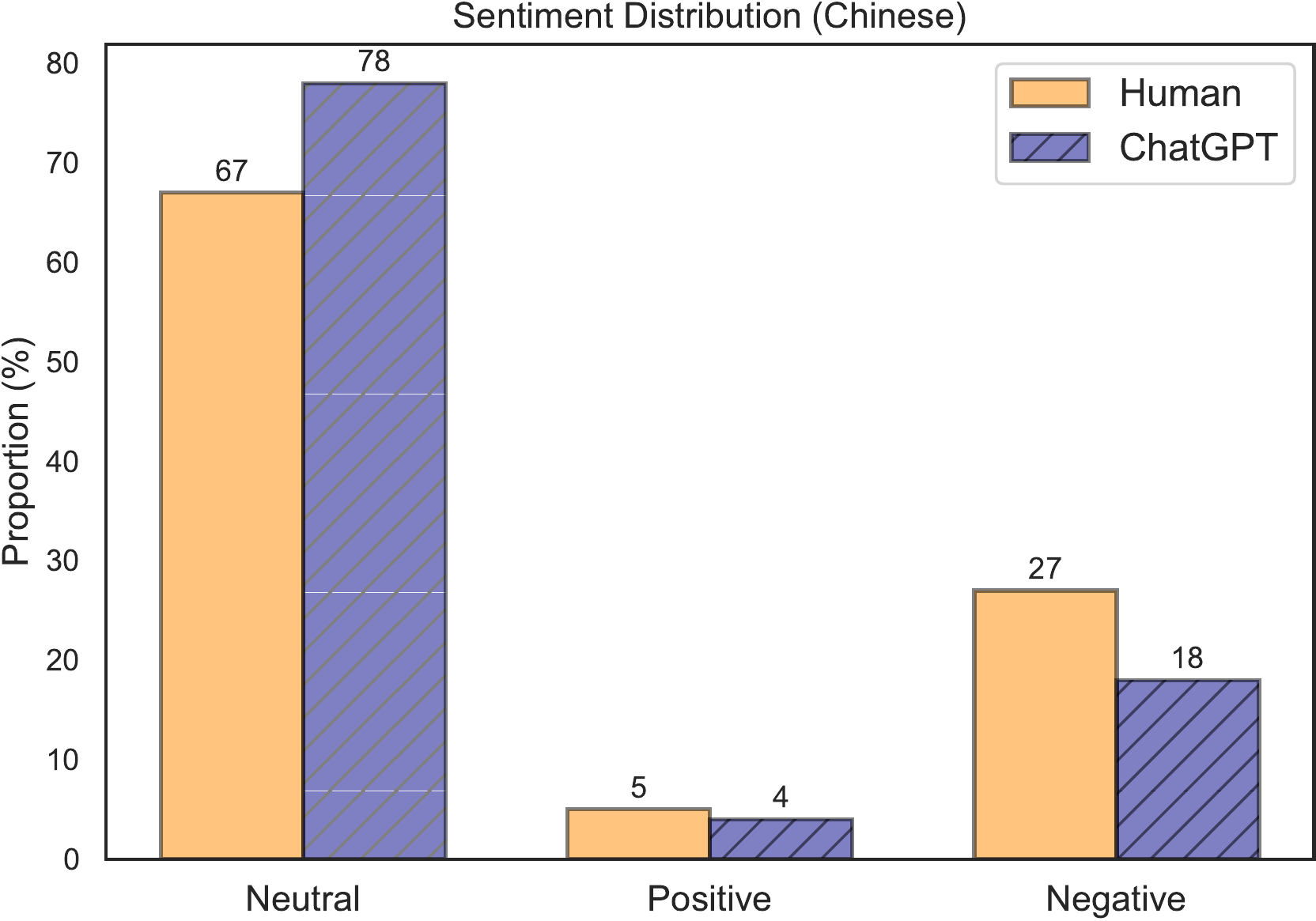}
         \caption{Sentiment distribution of the HC3-Chinese}
         \label{fig:zh-sentiment}
     \end{subfigure}
 \caption{Proportions of three kinds of sentiments (neutral, positive, and negative) in our corpus.}
 \label{fig:sentiment}
\end{figure*}

Figure \ref{fig:sentiment} shows the comparison of the sentiment distribution of humans and ChatGPT. Several findings can be drawn from the results: First, we find that the proportion of neutral emotions is the largest for both humans and ChatGPT, which is in line with our expectations. However, \textbf{ChatGPT generally expresses more neutral sentiments than humans}. Then, the proportion of negative emotions is significantly higher than that of positive emotions. Notably, \textbf{humans express significantly more negative emotions than ChatGPT}. The proportion of humans' positive emotions is also slightly higher than that of ChatGPT. Overall, ChatGPT is less emotional than humans, though it is not completely emotionless. 

\begin{figure*}[t]
\centering
\begin{subfigure}[b]{0.245\textwidth}
\centering
\includegraphics[width=\textwidth]{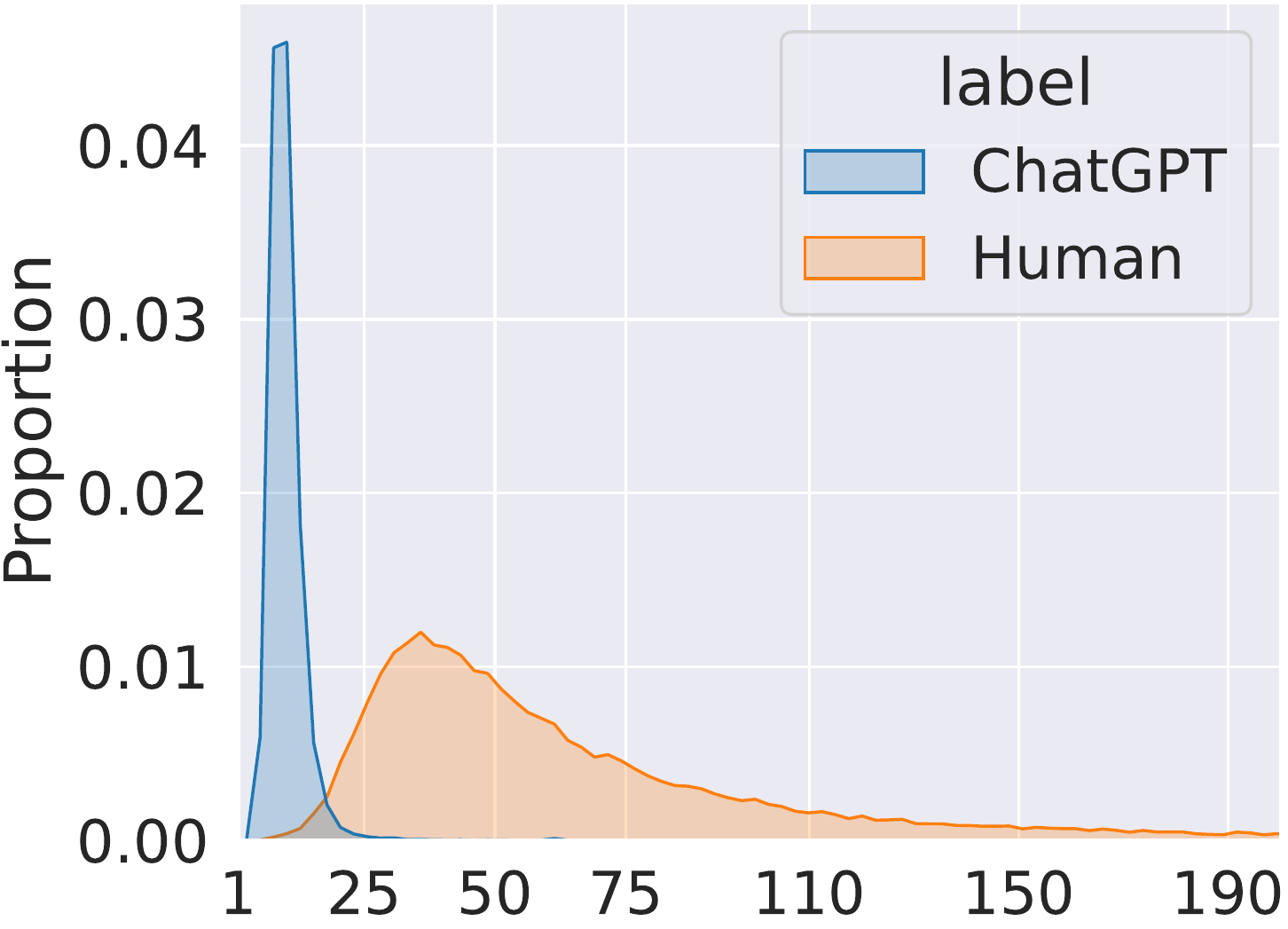}
\caption{English text ppl}
\label{fig:en-text-ppl}
\end{subfigure}
\begin{subfigure}[b]{0.245\textwidth}
\centering
\includegraphics[width=\textwidth]{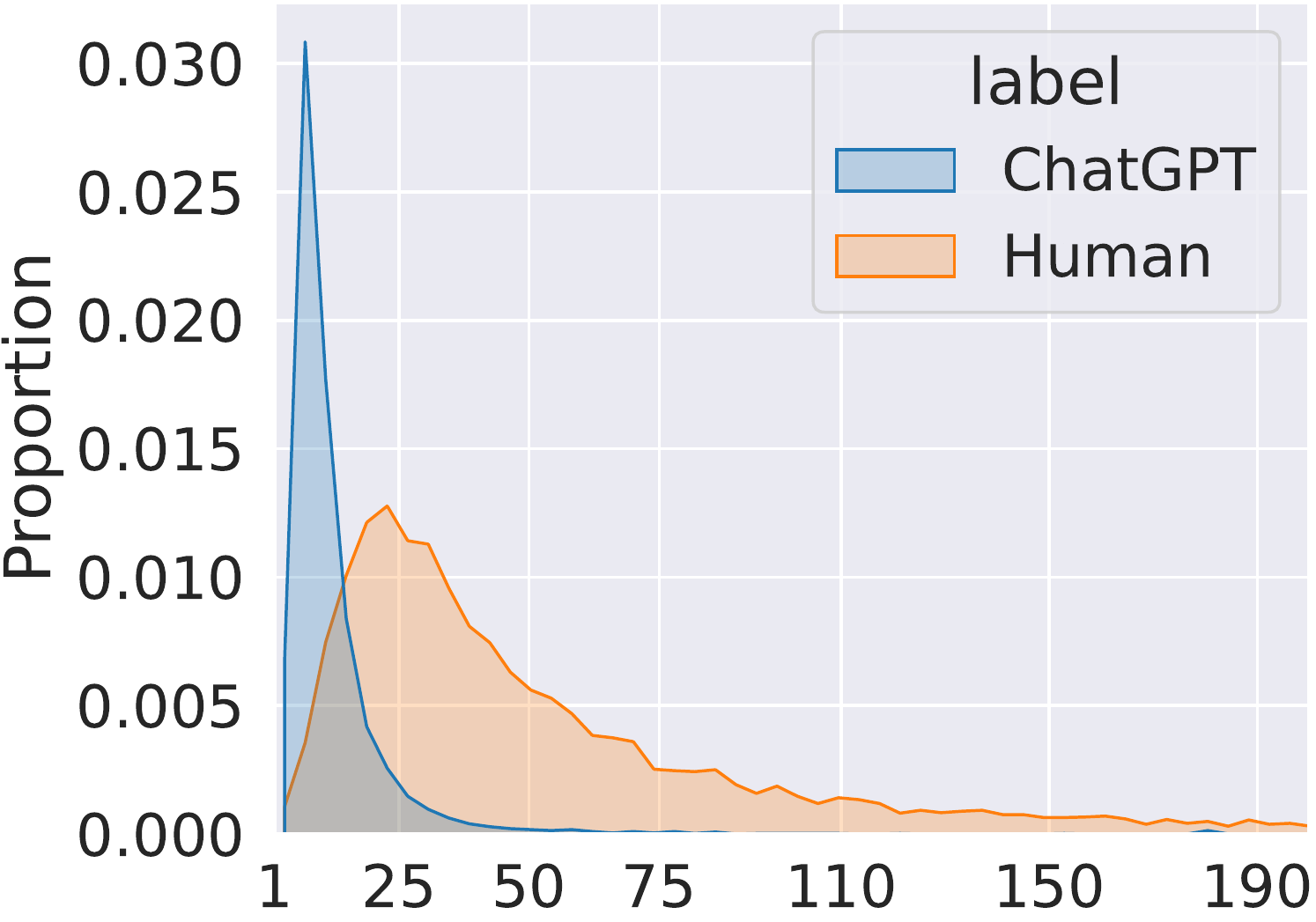}
\caption{English sent ppl}
\label{fig:en-sent-ppl}
\end{subfigure}
\begin{subfigure}[b]{0.245\textwidth}
\centering
\includegraphics[width=\textwidth]{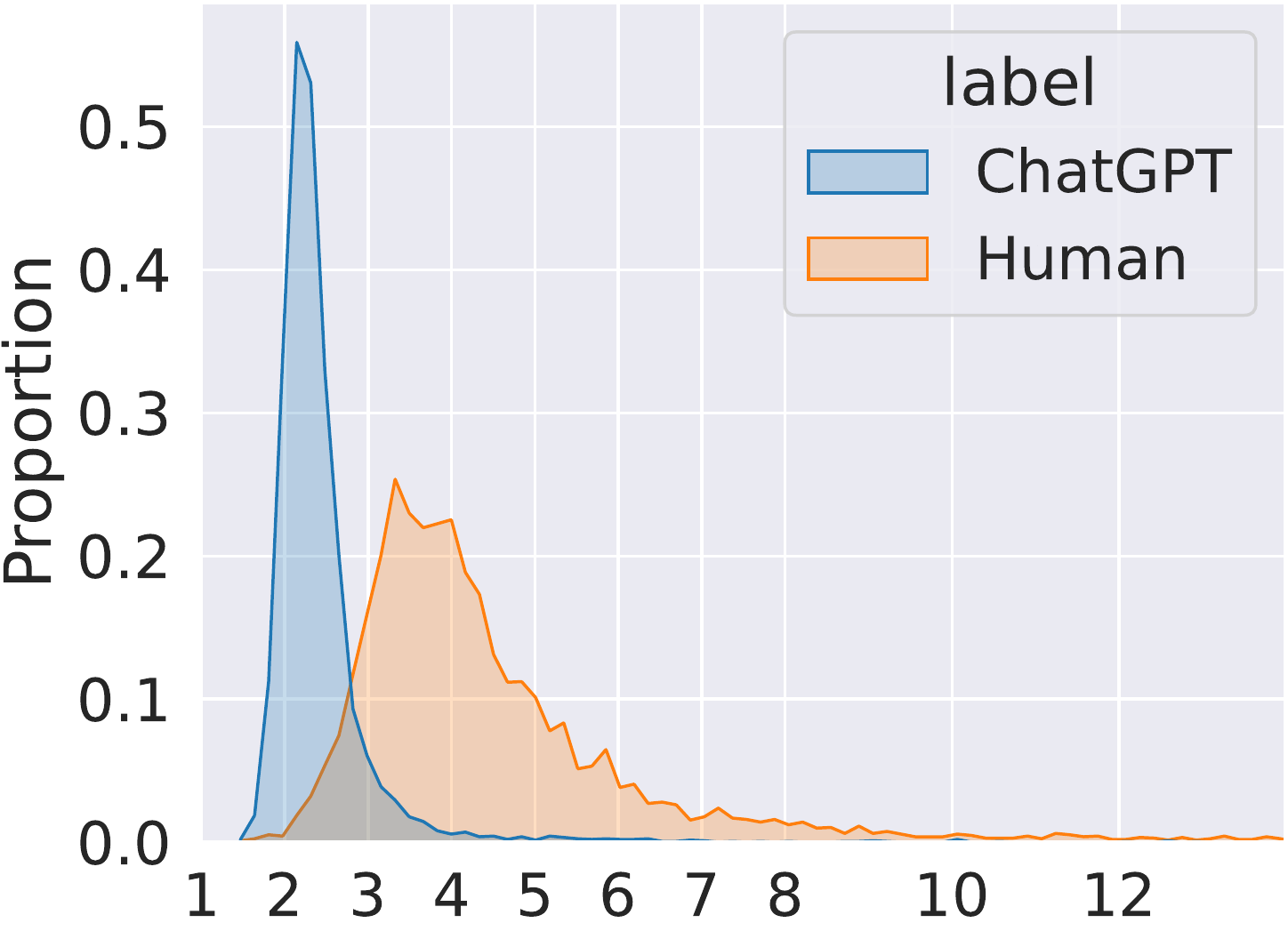}
\caption{Chinese text ppl}
\label{fig:zh-text-ppl}
\end{subfigure}
\begin{subfigure}[b]{0.245\textwidth}
\centering
\includegraphics[width=\textwidth]{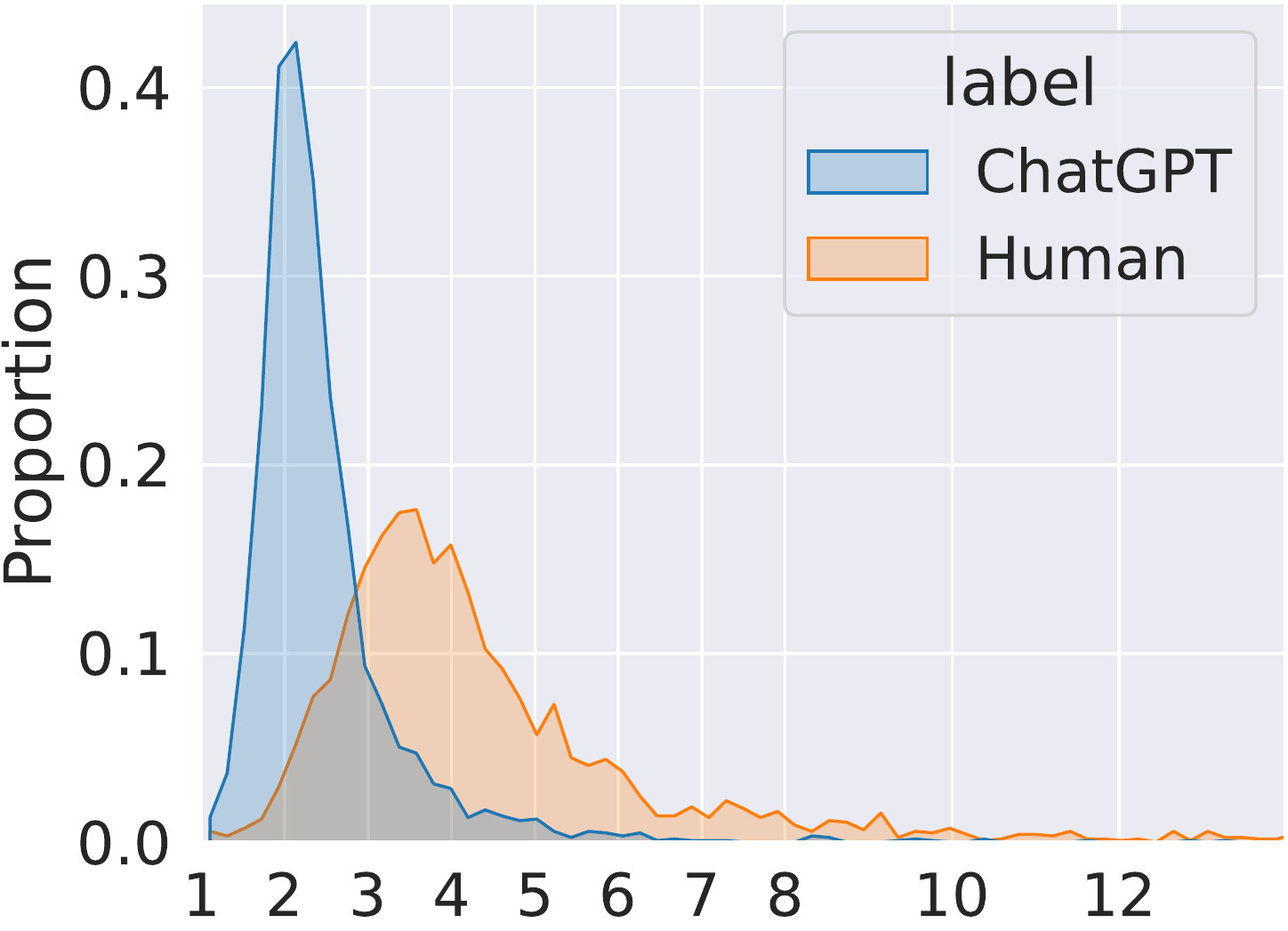}
\caption{Chinese sent ppl}
\label{fig:zh-sent-ppl}
\end{subfigure}
\caption{
PPL distributions on both English and Chinese data, as well as both text and sentence levels.
}
\label{fig:ling-ppl}
\end{figure*}

\subsection{Language Model Perplexity}
\label{sec:ppl}

The perplexity (PPL) is commonly used as a metric for evaluating the performance of language models (LM).
It is defined as the exponential of the negative average log-likelihood of the text under the LM.
A lower PPL indicates that the language model is more confident in its predictions, and is therefore considered to be a better model.
The training of LMs is carried out on large-scale text corpora, it can be considered that it has learned some common language patterns and text structures.
Therefore, we can use PPL to measure how well a text conforms to common characteristics.

We use the open-source GPT-2 small\footnote{\url{https://huggingface.co/gpt2}} (Wenzhong-GPT2-110M\footnote{\url{https://huggingface.co/IDEA-CCNL/Wenzhong-GPT2-110M}} for Chinese) model to compute the PPL (both text-level and sentence-level\footnote{
For English text, we used NLTK\cite{Bird_Natural_Language_Processing_2009} for sentence segmentation (HarvestText for Chinese).
} PPLs) of the collected texts.
The PPL distributions of text written by humans and text generated by ChatGPT are shown in Figure \ref{fig:ling-ppl}.

It is clearly observed that, regardless of whether it is at the text level or the sentence level, the content generated by ChatGPT has relatively lower PPLs compared to the text written by humans.
ChatGPT captured common patterns and structures in the text it was trained on, and is very good at reproducing them.
As a result, text generated by ChatGPT have relatively concentrated low PPLs.

Humans have the ability to express themselves in a wide variety of ways, depending on the context, audience, and purpose of the text they are writing.
This can include using creative or imaginative elements, such as metaphors, similes, and unique word choices, which can make it more difficult for GPT2 to predict.
Therefore, human-written texts have more high-PPL values, and show a long-tailed distribution, as demonstrated in Figure \ref{fig:ling-ppl}.

\section{ChatGPT Content Detection}
AI-generated content (AIGC) is becoming increasingly prevalent on the internet, and it can be difficult to distinguish it from human-generated content, as shown in our human evaluation (sec \ref{sec:human}). Therefore, AIGC detectors are needed to help identify and flag content that has been created by a machine, to reduce the potential risks to society caused by improper or malicious use of AI models, and to improve the transparency and accountability of the information that is shared online.

In this section, we conduct several empirical experiments to investigate the ChatGPT content detection systems. Detecting AI-generated content is a widely studied topic \cite{jawahar-etal-2020-automatic,pu2023deepfake}. Based on these \cite{solaiman2019release,gehrmann-etal-2019-gltr,pu2023deepfake}, we establish three different types of detection systems, including machine learning-based and deep learning-based methods, and evaluate them on different granularities and data sources. Detailed results and discussions are provided.

\subsection{Methods}
Detection of machine-generated text has been gaining popularity as text generation models have advanced in recent years\cite{jawahar-etal-2020-automatic,pu2023deepfake}.
Here, we implement three representative methods from classic machine learning and deep learning, i.e, a logistic regression model trained on the GLTR Test-2\cite{gehrmann-etal-2019-gltr} features,  a deep classifier for single-text detection and a deep classifier for QA detection. The deep classifiers for both single-text and QA are based on RoBERTa \cite{liu-et-al-roberta}, a strong pre-trained Transformer \cite{vaswani2017attention} model. In fact, algorithms for OOD detection or anomaly detection \cite{han2022adbench} can also be applied to develop ChatGPT content detectors, which we leave for future work.

\paragraph{GLTR.} \cite{gehrmann-etal-2019-gltr} studied three tests to compute features of an input text.
Their major assumption is that to generate fluent and natural-looking text, most decoding strategies sample high probabilities tokens from the head of the distribution.
We select the most powerful Test-2 feature, which is the number of tokens in the Top-10, Top-100, Top-1000, and 1000+ ranks from the LM predicted probability distributions.
And then a logistic regression model is trained to finish the classification.

\paragraph{RoBERTa-\textit{sinlge}.} A deep classifier based on the pre-trained LM is always a good choice for this kind of text classification problem.
It is also investigated in many studies and demo systems \cite{solaiman2019release,fagni2021tweepfake,pu2023deepfake}.
Here we fine-tune the RoBERTa \cite{liu-et-al-roberta} model.

\paragraph{RoBERTa-\textit{QA}.} While most content detectors are developed to classify whether a single piece of text is AI-generated, we claim that a detector that supports inputting both a question and an answer can be quite useful, especially for question-answering scenarios. Therefore, we decide to also build a QA version detector. The RoBERTa model supports a text pair input format, where a separating token is used to join a question and its corresponding answer.


\subsection{Implementation Details}

For the LM used by GLTR, we use gpt2-small \cite{radford2019gpt2} for English, and Wenzhong-GPT2-110M released by \cite{fengshenbang} for Chinese, it is the same with sec. \ref{sec:ppl}.  
For RoBERTa-based deep classifiers, we use \texttt{roberta-base}\footnote{\url{https://huggingface.co/roberta-base}} and \texttt{hfl/chinese-roberta-wwm-ext}\footnote{\url{https://huggingface.co/hfl/chinese-roberta-wwm-ext}} checkpoints for English and Chinese, respectively. All the above models are obtained from huggingface \texttt{transformers} \cite{wolf-etal-2020-transformers}.

 We train the logistic regression model by sklearn \cite{scikit-learn} on the GLTR Test-2 features from trainset, and search hyper-params following the code of \cite{pu2023deepfake}.
 The RoBERTa-based detectors are trained by the facilities of \texttt{transformers}.
 Specifically, we use the AdamW optimizer, setting batch size to 32 and learning rate to $5e-5$.
 We finetune models by 1 epoch for English, and 2 epochs for Chinese.

\begin{figure}[t]
    \centering
    \includegraphics[width=0.98\textwidth]{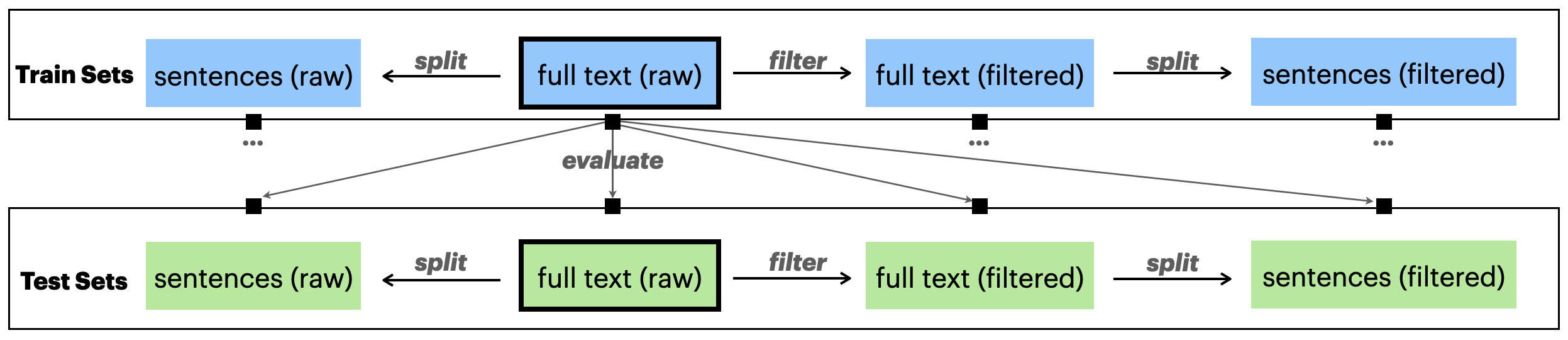}
    \caption{The experiment design for the training and testing of detectors. Different dataset versions are generated through filtering or splitting.}
    \label{fig:exp-design}
\end{figure}

\subsection{Experiment Design}
The HC3 dataset consists of questions and their corresponding human/ChatGPT answers. We extracted all the \texttt{<question, answer>} pairs, and assigned label \texttt{0} to pairs with human answers and label \texttt{1} to pairs with ChatGPT answers.

Simply using the original answers from humans and ChatGPT to train a binary classifier is the most straightforward way. However, there might be some issues by doing so:
\begin{itemize}
    \item First, based on the observations in Section \ref{sec:human-eval-sum}, both human answers and ChatGPT answers may contain some obvious indicating words that may influence the effectiveness of models;
    \item Second, users may want to detect whether a single sentence is generated by ChatGPT, instead of the full text. This can be quite difficult for a classifier that is only trained on full texts;
    \item Third, taking the corresponding question of the answer into account may help the detector to make a more accurate judgment, compared with only considering the answer itself. This can be widely applied to many QA platforms (like Quora, Stack Overflow, and Zhihu) to find out which answer below a certain question is generated by AI.
\end{itemize}

Therefore, we design different groups of experiments to study these key questions:\\
$\bullet$ How will the indicating words influence the detector?\\
$\bullet$ Is it more challenging for the ChatGPT detectors to detect sentence-level content? Is it harder to train a sentence-level classifier?\\
$\bullet$ Can the corresponding question help detectors detect the origin of the answer more accurately?

Figure \ref{fig:exp-design} shows how we generate different types of training and testing sets. Specifically, we use the collected raw corpus to construct the first train-test sets (the "full text (raw)" in the figure), which we call the \textbf{\textit{raw-full}} version. Then we filter away the indicated words in the text to obtain the \textbf{\textit{filtered-full}} version. By splitting the full text into sentences, we obtain the \textbf{\textit{raw-sent}} version and the \textbf{\textit{filtered-sent}} version. We also combine the full text and the sentences into a mixed version, namely the \textbf{\textit{raw-mix}} and \textbf{\textit{filtered-mix}} version. Overall, we have six different versions of training and testing sets.
Evaluating a model's performance on version B's testing set which is trained on version A's training set can be seen as an out-of-distribution (OOD) generalization evaluation, which is more challenging since it requires the model to be robust when facing sample style changes.

\subsection{Results}

Following the above experiment design, we conduct comprehensive empirical studies on all kinds of derived corpus.
Table \ref{tab:result-main} shows the test F1 scores.

\input{tables/main_table.tex}

\subsubsection{Which detector(s) is more useful? ML-based or DL-based? and Why? }

According to Table \ref{tab:result-main}, we can derive following conclusions:

Firstly, \textbf{the robustness of RoBERTa-based-detector is better than GLTR}. The F1-scores of RoBERTa decrease slightly (1.5-2\% in English datasets and 2-3\% in Chinese datasets) when sentences are split by comparing the leading diagonal elements in \textit{raw}$\rightarrow$\textit{raw} and \textit{filtered}$\rightarrow$\textit{filtered}. In contrast, the GLTR reduces significantly by over 10\% in English datasets, and above 15\% in Chinese datasets. Above all, the RoBERTa-based-detector is more robust with anti-interference character. In contrast, the GLTR reduces significantly by over 10\% in English datasets, above 15\% in Chinese datasets. Above all, the RoBERTa-based-detector is more robust with anti-interference character.

Secondly, \textbf{RoBERTa-based-detector is not affected by indicating words.} The F1-scores of RoBERTa only slightly decreased by 0.03\% in English \textit{full} dataset, and 0.65\% in Chinese \textit{full} dataset, as seen in the minus of relevant leading diagonal elements in \textit{raw}$\rightarrow$\textit{raw} versus \textit{filtered}$\rightarrow$\textit{filtered}. On the contrary, evaluations based on GLTR decrease by up to 3.1\% on Chinese datasets, though tiny rise on English datasets, indicating that GLTR is sensitive to indicating words, easily influenced by the patterns of ChatGPT.

Lastly, \textbf{RoBERTa-based-detector is effective in handling Out-Of-Distribution scenarios.} When compared to the original model, it demonstrates a significant decrease in performance on GLTR's OOD test datasets, with a drop of up to 28.8\% on English datasets(\textit{filtered-full}$\rightarrow$\textit{filtered-full} $-$ \textit{filtered-full}$\rightarrow$\textit{filtered-sent}) and 45.5\% on Chinese datasets(\textit{raw-full}$\rightarrow$\textit{raw-full} $-$ \textit{raw-full}$\rightarrow$\textit{raw-sent}). However, RoBERTa maintains consistent performance with F1-scores varying by no more than 19\%.

\subsubsection{How will the indicating words influence the detector?}
We first collected a bunch of indicating words for both humans and ChatGPT. For example, ChatGPT's indicating words (or phrases) include "AI assistant", "I'm sorry to hear that", and "There're a few steps...", etc. and humans' indicating words may include "Hmm", "Nope", "My view is", etc. In the filtered version, we remove all sentences in the answers that contain the indicating words for both humans and ChatGPT.

According to Table \ref{tab:result-main}, \textbf{removing the indicating words helps the models trained on full-text to perform better across different content granularities}. For example, the RoBERTa-\textit{filter-full} performs significantly better than RoBERTa-\textit{raw-full} in terms of sentence-level and mix-level evaluations, improving more than 3\% F1 scores on average.
However, \textbf{the filtering may slightly hurt the performances of the models trained on sentences.} This may be because the indicating words play a bigger part in the sentence-level text compared with the full text. Removing the indicating words may make some sentences literally unable to be distinguished.

\subsubsection{Which granularity is more difficult to detect? Full-text or sentence?}
Through the extensive experimental results in Table~\ref{tab:result-full-sent-mix}, we conclude that \textbf{detecting ChatGPT generated texts is more difficult in a single sentence than in a full text}. This conclusion can be proved by the following two points: 
First, our results show that both English and Chinese sentence-based detectors (i.e., \textit{raw-sent} and \textit{filtered-sent} versions) achieve satisfactory results w.r.t. the testing task of detecting either ChatGPT generated paragraphs or sentences, whereas the opposite is not true——\textit{raw-full} and \textit{filtered-full} are relatively inferior when detecting ChatGPT generated sentences. In other words, detectors trained on "hard samples" (i.e., sentence corpus) are much easier to solve simple task (i.e., detecting full corpus), while "simple samples" (i.e., full corpus) may be less useful for solving more difficult task (i.e., sentence corpus).

Second, we observe that although both full and sentence corpus are provided in the \textit{raw-mix} and \textit{filtered-mix} versions, it is still more difficult for them to detect single sentences generated by ChatGPT. This is even more obvious for the Chinese corpus, where the F1-score of \textit{raw-mix} trained on the Chinese corpus is 94.09\% for testing raw sentence answers, compared to that 97.43\% for testing raw full answers. Similar results can be observed for the filtered corpus, where F1-score of \textit{filtered-mix} is 95.61\% for testing filtered sentence answers, compared to its F1-score of 97.66\% for testing filtered full answers.
One possible explanation is that the expression pattern of ChatGPT is more obvious (therefore more easily detected) when paragraphs of text are provided, whereas it is more difficult to detect generated single sentences.

\input{tables/full_sent_mix_table.tex}

\subsubsection{Which corpus is more helpful for model training? Full-text, sentence, or mix of the two?}
We find that both English and Chinese RoBERTa-based \textbf{detectors are more robust when fine-grained corpus data is available in model training}. The sentence-based detectors outperform full-based detectors w.r.t. F1-scores, while the latter can be significantly improved when the sentence corpus is injected in model training, as we observe that mix-based detectors also achieve satisfactory results. 
For English corpus, \textit{raw-full} only achieves 81.89\% F1-score for testing sentence answers, while \textit{raw-sent} is significantly better with 98.43\% F1-score, as shown in Table~\ref{tab:result-full-sent-mix}. Moreover, the relatively inferior detection performance can be improved by injecting sentence answers into the detector, where we find that \textit{raw-mix} can also obtain significant improvement (with 98.31\% F1-score) over the detectors trained on only full answers. Similar conclusions can be acquired for the filtered versions, where both \textit{filtered-sent} and \textit{filtered-mix} significantly outperform \textit{filtered-full} version w.r.t. F1-score, which holds for both English and Chinese corpus.

We indicate that the above conclusions could also hold for other types of detectors like GLTR Test-2 feature-based detectors, as is shown in Table~\ref{tab:result-main}. For GLTR Test-2, the average performance of F1-score of \textit{raw-full} and \textit{filtered-full} is 61.74\% and 69.47\%, respectively, compared to that of \textit{raw-sent} 76.26\% and \textit{filtered-sent} 76.41\%, where the performance of detectors trained on the mixed corpus is close to the sentence-based versions. 

Taking into account the conclusions of the previous paragraph about the detection difficulty between full and sentence answers, we indicate that the fine-grained corpus is helpful for distinguishing ChatGPT generated texts, as it additionally provides guidance and hints in model training for detecting the subtle patterns of ChatGPT hidden in single sentences.

\subsubsection{Will a QA-style detector be more effective than a single-text detector?} 
Table \ref{tab:result-qa} demonstrates the results of both \textit{raw-full} and \textit{filtered-full} models across all test datasets.

On English datasets, the QA model's F1-scores are superior to that of the single model, except for two \textit{full} test datasets, where it averages 97.48\% F1-scores and surpasses single model by 5.63\%. There exist some differences in Chinese datasets, where the single model outperforms QA in \textit{raw-full} train dataset. However, the QA model still yields the best evaluation at 94.22\%.

In conclusion, \textbf{the QA model is generally more effective than the single model and is suitable for filtered scenarios.
And the QA training makes models more robust to the sentence inputs.}

\input{tables/qa_table.tex}

\subsubsection{Which data sources are more difficult for the ChatGPT detectors? and What are the conditions that make it easier to detect ChatGPT?}

As shown in Table \ref{tab:result-source}, the evaluation results based on \textit{filtered-full} model are separated by various sources in our HC3 dataset.

On the English datasets, the F1-scores for human answers are slightly higher than those for ChatGPT without any exceptions, regardless of whether RoBERTa or GLTR is used on full-text test datasets. However, the F1-scores for ChatGPT are highly inconsistent on transferring test datasets particularly \texttt{open-qa} dataset with varying performance. \textbf{In terms of data resource, \texttt{reddit-eli5} and \texttt{finance-en} has higher values, while \texttt{wiki-csai} poses a challenge for detectors.}

On the Chinese datasets, the F1-scores of humans and ChatGPT are comparable with no significant difference. This suggests that the difficulty in detecting ChatGPT depends on the data source. \textbf{It is observed that \texttt{open-qa} and \texttt{baike} have better performance, whereas the \texttt{nlpcc-dbqa} has lower performance. }

Above all, the evaluations on Chinese dataset show more stability on transferring test dataset compared to the English datasets. Furthermore, it's evident that the F1-scores of ChatGPT are lower than those of human answers, regardless of whether the dataset is English or Chinese. This indicates that \textbf{ChatGPT's detector relies more heavily on In-Distribution models.}

\input{tables/source_table.tex}




\section{Conclusion}
In this work, we propose the HC3 (Human ChatGPT Comparison Corpus) dataset, which consists of nearly 40K questions and their corresponding human/ChatGPT answers. Based on the HC3 dataset, we conduct extensive studies including human evaluations, linguistic analysis, and content detection experiments. The human evaluations and linguistics analysis provide us insights into the implicit differences between humans and ChatGPT, which motivate our thoughts on LLMs' future directions. The ChatGPT content detection experiments illustrate some important conclusions that can provide beneficial guides to the research and development of AIGC-detection tools. We make all our data, code, and models publicly available to facilitate related research and applications at \url{https://github.com/Hello-SimpleAI/chatgpt-comparison-detection}.

\section{Limitations}
Despite our comprehensive analysis of ChatGPT, there are still several limitations in the current paper, which will be considered for improvement in our future work:
\begin{itemize}
    \item[1.] Despite our efforts in data collection, the amount and range of collected data are still not enough and the data from different sources are unbalanced, due to limited time and resources. To make more accurate linguistic analyses and content detection, more data with different styles, sources, and languages are needed;
    \item [2.] Currently, all the collected ChatGPT answers are generated \textbf{without special prompts}. Therefore, the analysis and conclusions in this paper are built upon ChatGPT's most general style/state. For example, using special prompts such as "Pretending you are Shakespeare..." can generate content that bypasses our detectors or make the conclusions in this paper untenable;
    \item [3.] ChatGPT (perhaps) is mainly trained on English corpus while less on Chinese. Therefore, the conclusions drawn from the HC3-Chinese dataset may not always be precise.
    
\end{itemize}

\section*{Acknowledgments}
We would like to thank the volunteers that participated in our human evaluations, many of them are our good friends and dear family members.
We would like to thank Junhui Zhu (BLCU-ICALL) for the valuable discussions on linguistic analysis. Biyang Guo would like to thank Prof. Hailiang Huang and Prof. Songqiao Han (AI Lab, SUFE) for providing insightful feedback on the topics and directions for this project.
Xin Zhang would like to thank Yu Zhao (NeXt, NUS and CIC, TJU) for sharing the OpenAI account.
Finally, we thank all team members of this project for their unique contributions. We together make this possible.




\clearpage
\bibliography{ref}
\bibliographystyle{plain}

\appendix

\section{Appendix}

\subsection{HC3 Dataset Splits Creation} \label{app:splits}

We create 5 and 7 splits for HC3 English and Chinese, respectively.
Most of the data come from the publicly available Question-Answering (QA) datasets, where details are listed in the following.
For these QA data, we directly input the questions to ChatGPT and collect at least one answer.

We also crawled some wiki concepts and explanations from Wikipedia and BaiduBaike, where explanations are treated as human expert answers and concepts are used to construct the questions, details ref to bellow paragraphs.

For HC3-English, we create five dataset splits:
\begin{itemize}
    \item[1.] \texttt{reddit\_eli5}. Sampled from the ELI5 dataset \cite{reddit-eli5_lfqa}.
    \item[2.] \texttt{open\_qa}. Sampled from the WikiQA dataset \cite{yang2015wikiqa}.
    \item[3.] \texttt{wiki\_csai}. We collected the descriptions of hundreds of computer science-related concepts from Wikipedia\footnote{https://www.wikipedia.org/} as the human experts' answers to questions like "Please explain what is \texttt{<concept>}?"
    \item[4.] \texttt{medicine}. Sampled from the Medical Dialog dataset \cite{chen2020MedDialog-en-zh}.
    \item[5.] \texttt{finance}. Sampled from the FiQA dataset \cite{fiqa-2018}, which is built by crawling StackExchange\footnote{https://stackexchange.com/} posts under the Investment topic.
\end{itemize}

For HC3-Chinese, we create seven dataset splits:
\begin{itemize}
    \item[1.] \texttt{open\_qa}. Sampled from the WebTextQA and BaikeQA corpus in \cite{chinese_corpus-webtext}.
    \item[2.] \texttt{baike}. We collected the descriptions of more than a thousand information science-related concepts from BaiduBaike\footnote{https://baike.baidu.com/} as the human experts' answers to questions like "\begin{CJK*}{UTF8}{gbsn}我有一个计算机相关的问题，请用中文回答，什么是\end{CJK*}\texttt{<concept>}" 
    \item[3.] \texttt{nlpcc\_dbqa}. Sampled from the NLPCC-DBQA dataset \cite{duan2016nlpcc}.
    \item[4.] \texttt{medicine}. Sampled from the Medical Dialog dataset \cite{chen2020MedDialog-en-zh}.
    \item[5.] \texttt{finance}. Sampled from the FinanceZhidao dataset \cite{SophonPlus2019financezhidao}.
    \item[6.] \texttt{psychology} Sampled from a public Chinese Psychological Question Answering Dataset\footnote{https://aistudio.baidu.com/aistudio/datasetdetail/38489}.
    \item[7.] \texttt{law}. Sampled from the LegalQA dataset\footnote{https://github.com/siatnlp/LegalQA}.
\end{itemize}

\subsection{Additional Results} 
\label{app: depency}
\begin{figure}[htp]
	\centering
	\begin{minipage}[t]{\textwidth}
		\centering
		\includegraphics[width=\textwidth]{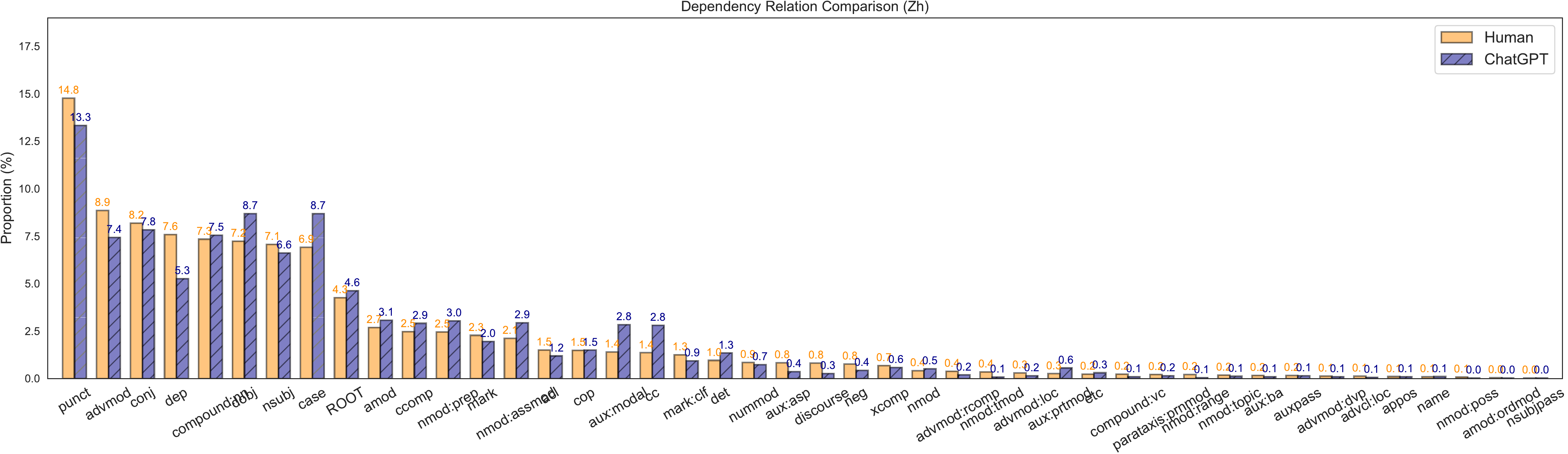}
	\end{minipage}
        \vspace{0.15in}
	\begin{minipage}[t]{\textwidth}
		\centering
		\includegraphics[width=\textwidth]{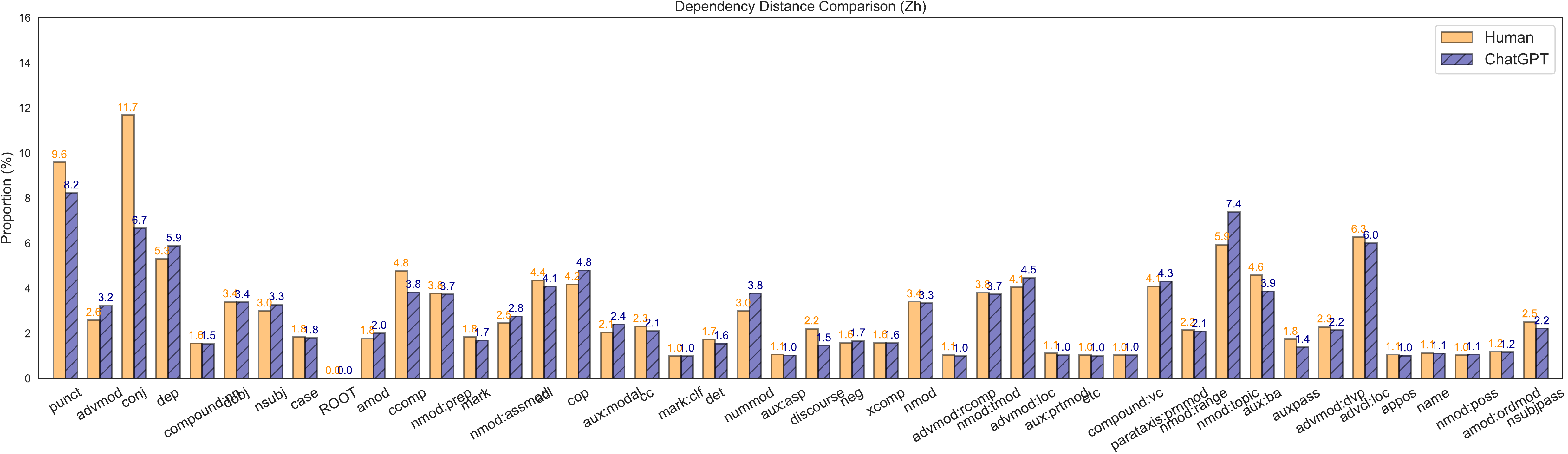}
	\end{minipage}
	\caption{Top-30 dependency relations  (upper) and corresponding dependency distances (lower) comparison between human and ChatGPT answers in the HC3-Chinese. Results are sorted by relations proportion of human answers.}
        \label{fig:dep-and-dist-yuxuan-appendix}
\end{figure}

Here we demonstrate the additional results of dependency relations for the Chinese corpus, as is shown in Figure~\ref{fig:dep-and-dist-yuxuan-appendix}. The conclusion is basically consistent with the main paper.

Other detailed results, including vocabulary features, sentiment analyses, and dependency parsing results for each data source are all available at our project GitHub repository at \url{https://github.com/Hello-SimpleAI/chatgpt-comparison-detection}.

\subsection{Human Evaluations Examples}
For evaluation examples of our human evaluations, please visit our project GitHub repository at \url{https://github.com/Hello-SimpleAI/chatgpt-comparison-detection}.

\end{document}

%% file: tables/dataset_meta-biyang.tex
\begin{table}[t]
\centering
\resizebox{0.98\textwidth}{!}{
\begin{tabular}{lcccl}

\multicolumn{5}{l}{\textbf{HC3-English }}\\
\toprule[1.5pt]

 & \textbf{\# Questions} & \textbf{\# Human Answers} & \textbf{\# ChatGPT Answers}  & \bf Source \\
\midrule
\textbf{All}             & 24322  & 58546      & 26903  \\
\midrule
\textit{reddit\_eli5}    & 17112  & 51336      & 16660    &  ELI5 dataset \cite{reddit-eli5_lfqa} \\
\textit{open\_qa} & 1187   & 1187       & 3561     &  WikiQA dataset \cite{yang2015wikiqa} \\
\textit{wiki\_csai} & 842    & 842        & 842      &  Crawled Wikipedia (\ref{app:splits}) \\
\textit{medicine}        & 1248   & 1248       & 1337     &  Medical Dialog dataset \cite{chen2020MedDialog-en-zh} \\
\textit{finance}         & 3933   & 3933       & 4503     &  FiQA dataset \cite{fiqa-2018} \\
\bottomrule
\\

\multicolumn{5}{l}{\textbf{HC3-Chinese }}\\
\toprule[1.5pt]

 & \textbf{\# Questions} & \textbf{\# Human Answers} & \textbf{\# ChatGPT Answers}  & \bf Source \\
\midrule
\textbf{All}             & 12853   & 22259     & 17522  \\
\midrule
\textit{open\_qa} & 3293   & 7377       & 3991    &  WebTextQA \& BaikeQA \cite{chinese_corpus-webtext}\\
\textit{baike}           & 4617   & 4617       & 4617    &  Crawled BaiduBaike (\ref{app:splits}) \\
\textit{nlpcc\_dbqa}     & 1709   & 1709       & 4253    &  NLPCC-DBQA dataset \cite{duan2016nlpcc} \\
\textit{medicine}        & 1074   & 1074       & 1074    &  Medical Dialog dataset \cite{chen2020MedDialog-en-zh} \\
\textit{finance}         & 689    & 1572       & 1983    &  ChineseNlpCorpus (\ref{app:splits})\\
\textit{psychology}      & 1099   & 5220       & 1099    &  from Baidu AI Studio (\ref{app:splits}) \\
\textit{law}             & 372    & 690        & 505     &  LegalQA dataset (\ref{app:splits}) \\
\bottomrule\\
\end{tabular}
}
\caption{Meta-information of the HC3 dataset. The English (resp. Chinese) contains 5 (resp. 7) splits.}
\label{dataset_meta-biyang}
\end{table}

%% file: tables/human_evaluation.tex
\begin{table}[t]
\centering
\resizebox{0.85\textwidth}{!}{
\begin{tabular}{lcccc}
\multicolumn{5}{l}{\textbf{Human Evaluation (En) }}\\
\toprule[1.5pt]
  & \textbf{Pair-expert} & \textbf{Single-expert} & \textbf{Single-amateur} & \textbf{Helpfulness} \\ \midrule
   \textbf{All}    &  0.90   &  0.81   & 0.48  &   0.57 \\ \midrule
\textit{reddit\_eli5}       &   0.97      &   0.94       & 0.57 &  0.59   \\
\textit{open\_qa} & 0.98    & 0.78    & 0.34     & 0.72   \\ 
\textit{wiki\_csai} & 0.97       & 0.61     & 0.39     & 0.71   \\ 
\textit{medical} & 0.97       & 0.97     & 0.50   & 0.23  \\
\textit{finance} & 0.79      & 0.73      & 0.58      & 0.60    \\ 

\bottomrule
\\
\multicolumn{5}{l}{\textbf{Human Evaluation (Zh) }}\\
\toprule[1.5pt]
    & \textbf{Pair-expert} & \textbf{Single-expert} & \textbf{Single-amateur} & \textbf{Helpfulness} \\ \midrule
\textbf{All} & 0.93   &  0.86  &  0.54  &   0.54 \\ \midrule
\textit{open\_qa} &  1.00  & 0.92   &  0.47 &  0.50  \\
\textit{baike}      &    0.76     & 0.64   & 0.60  & 0.60   \\ 
\textit{nlpcc\_dbqa} & 1.00       &  0.90                           & 0.13                                   & 0.63             \\ 
\textit{medicine} & 0.93       & 0.93                              & 0.57                                   & 0.30            \\
\textit{finance} & 0.86       &  0.84                            & 0.84                                   & 0.75             \\ 
\textit{psychology} & 1.00      & 1.00                              & 0.60                                  & 0.67             \\
\textit{law}  & 1.00      & 0.77                              & 0.56                                   & 0.56             \\ 

\bottomrule
\\
\end{tabular}
}
\caption{Human evaluations of ChatGPT generated answers for both English and Chinese.}
\label{human_eval_all}
\end{table}

%% file: tables/vocab_compare-biyang.tex
\begin{table}[t]
\centering
\resizebox{0.98\textwidth}{!}{
\begin{tabular}{l|lccc|lccc}
\toprule[1.5pt]
& \textbf{English} & \textbf{avg. len.} & \textbf{vocab size} & \textbf{density} &  \textbf{Chinese}  & \textbf{avg. len.} & \textbf{vocab size} & \textbf{density} \\ \midrule
human    &  \multirow{2}{*}{\textbf{All}}
& 142.50           & \textbf{79157}  & \textbf{2.33}   & \multirow{2}{*}{\textbf{All}} & 102.27         & \textbf{75483}  & \textbf{5.75}  \\
ChatGPT  &
& \textbf{198.14}  & 66622           & 1.41            &  & \textbf{115.3} & 45168           & 3.05 \\ \midrule

human    & \multirow{2}{*}{\textit{reddit\_eli5}}
& 134.21           & \textbf{55098}  & \textbf{2.46}   & \multirow{2}{*}{\emph{nlpcc\_dbqa}} & 24.44  & 10621  & \textbf{25.43} \\
ChatGPT  &
& \textbf{194.84}  & 44926           & 1.38            &  & \textbf{78.21}  & \textbf{11971}  & 8.96 \\ \midrule

human    & \multirow{2}{*}{\textit{open\_qa}}
& 35.09            & 9606            & \textbf{23.06} & \multirow{2}{*}{\textit{open\_qa}} & 93.68           & \textbf{40328}   & \textbf{13.13}    \\
ChatGPT  &
& \textbf{131.68}  & \textbf{16251}  & 10.40          &  & \textbf{150.66} & 26451            & 5.35  \\ \midrule

human    & \multirow{2}{*}{\textit{wiki\_csai}}
& \textbf{229.34}  & \textbf{15859}  & \textbf{8.21}  & \multirow{2}{*}{\emph{baike}}  & \textbf{112.25}  & \textbf{28966}  & \textbf{5.59}  \\
ChatGPT  &
& 208.33           & 9741            & 5.55           &  & 77.19      & 14041     & 3.94  \\ \midrule

human    &  \multirow{2}{*}{\textit{medicine}}
& 92.98            & \textbf{11847}  & \textbf{10.42} & \multirow{2}{*}{\textit{medicine}} & 92.34      & \textbf{9855}    & \textbf{9.94} \\
ChatGPT  &  
& \textbf{209.61}  & 7694            & 3.00           &  & \textbf{165.41}  & 7211     & 4.06  \\ \midrule

human    &  \multirow{2}{*}{\textit{finance}}
& 202.07           & \textbf{25500}  & \textbf{3.21}  & \multirow{2}{*}{\textit{finance}} & 80.76   & 2759  & \textbf{5.05}  \\
ChatGPT  &
& \textbf{226.01}  & 21411           & 2.41           &  & \textbf{120.84} & \textbf{4043}  & 4.94  \\  \midrule

human   & - & - & - & - &  \multirow{2}{*}{\textit{psychology}}  & \textbf{254.82}  & \textbf{16160}  & \textbf{5.77} \\
ChatGPT & - & - & - & - &                                        & 164.53      & 5897    & 3.26  \\ \midrule

human   & - & - & - & - &  \multirow{2}{*}{\textit{law}}  & 28.77   & 2093    & \textbf{19.55} \\
ChatGPT & - & - & - & - &                                 & \textbf{143.76}   & \textbf{3857}  & 7.21 \\
\bottomrule
\bottomrule[0pt]
\bottomrule[0pt]

\end{tabular}
}

\caption{Average answer length, vocabulary size and density comparisons on our corpus.}
\label{vocab_compare}
\end{table}

%% file: tables/main_table.tex
\begin{table}[h!]
\centering
\setlength{\tabcolsep}{5pt}
\resizebox{0.98\textwidth}{!}{
\begin{tabular}{cc|ccc|ccc|c|ccc|ccc|c}
\toprule[1.5pt]
&  & \multicolumn{7}{c|}{\bf English}   & \multicolumn{7}{c}{\bf Chinese}  \\
\midrule
\multicolumn{2}{c|}{\multirow{2}{*}{Test $\rightarrow$}}
& \multicolumn{3}{c|}{\textit{raw}}  & \multicolumn{3}{c|}{\textit{filtered}} & \multirow{2}{*}{Avg.} & \multicolumn{3}{c|}{\textit{raw}}   & \multicolumn{3}{c|}{\textit{filtered}} & \multirow{2}{*}{Avg.}  \\
\multicolumn{2}{c|}{}    & full & sent & mix                & full & sent & mix    &      & full & sent & mix          & full & sent & mix  \\
\midrule \midrule
\multicolumn{2}{c|}{Train $\downarrow$} & \multicolumn{14}{c}{\bf RoBERTa} \\ \midrule
\multirow{3}{*}{\textit{raw}}      & full & 99.82 & 81.89 & 84.67      & 99.72 & 81.00 & 84.07   & 88.53    & 98.79 & 83.64 & 86.32      & 98.57 & 82.77 & 85.85   & 89.32  \\
                                   & sent & 99.40 & 98.43 & 98.56      & 99.24 & 98.47 & 98.59   & \bf 98.78    & 97.76 & 95.75 & 96.11      & 97.68 & 95.31 & 95.77   & \bf 96.40  \\
                                   & mix  & 99.44 & 98.31 & 98.47      & 99.32 & 98.37 & 98.51   & 98.74    & 97.70 & 95.68 & 96.04      & 97.65 & 95.27 & 95.73   & 96.35  \\
\midrule
\multirow{3}{*}{\textit{filtered}} & full & 99.82 & 87.17 & 89.05      & 99.79 & 86.60 & 88.67   & 91.85    & 98.25 & 91.04 & 92.30      & 98.14 & 91.15 & 92.48   & 93.89  \\
                                   & sent & 96.97 & 97.22 & 97.19      & 99.09 & 98.43 & 98.53   & 97.91    & 96.60 & 92.81 & 93.47      & 97.94 & 95.86 & 96.26   & 95.49  \\
                                   & mix  & 96.28 & 96.43 & 96.41      & 99.45 & 98.37 & 98.53   & 97.58    & 97.43 & 94.09 & 94.68      & 97.66 & 95.61 & 96.01   & 95.91  \\
\midrule \midrule
\multicolumn{2}{c|}{Train $\downarrow$} & \multicolumn{14}{c}{\bf GLTR Test-2} \\ \midrule
\multirow{3}{*}{\textit{raw}}      & full & 98.26 & 71.58 & 76.15      & 98.22 & 70.19 & 75.23   & 81.61    & 89.61 & 44.02 & 53.72      & 85.89 & 43.58 & 53.62   & 61.74  \\
                                   & sent & 86.26 & 88.18 & 87.96      & 87.72 & 88.23 & 88.19   & 87.76    & 84.49 & 71.79 & 74.01      & 84.06 & 70.29 & 72.90   & 76.26  \\
                                   & mix  & 95.97 & 86.45 & 87.81      & 96.13 & 86.24 & 87.73   & 90.06    & 86.45 & 70.85 & 73.59      & 84.94 & 69.14 & 72.14   & 76.19  \\
\midrule
\multirow{3}{*}{\textit{filtered}} & full & 98.31 & 70.91 & 75.65      & 98.30 & 69.48 & 74.72   & 81.23    & 89.46 & 58.69 & 64.52      & 86.51 & 55.45 & 62.18   & 69.47  \\
                                   & sent & 84.00 & 88.25 & 87.71      & 85.68 & 88.35 & 87.99   & 87.00    & 84.56 & 71.85 & 74.07      & 84.22 & 70.59 & 73.18   & 76.41  \\
                                   & mix  & 95.36 & 86.73 & 87.97      & 95.60 & 86.56 & 87.92   & 90.02    & 86.30 & 71.00 & 73.70      & 84.98 & 69.45 & 72.40   & 76.31  \\
\bottomrule \\
\end{tabular}
}
\caption{F1 scores (\%) of different models on each testset, average of each language are reported.}
\label{tab:result-main}
\end{table}

%% file: tables/full_sent_mix_table.tex
\begin{table}[h!]
\centering
\setlength{\tabcolsep}{5pt}
\resizebox{0.98\textwidth}{!}{
\begin{tabular}{c|ccc|ccc|c|ccc|ccc|c}
\toprule[1.5pt]
 & \multicolumn{7}{c|}{\bf English}   & \multicolumn{7}{c}{\bf Chinese}  \\
\midrule
Test $\rightarrow$ & \multicolumn{3}{c|}{\emph{raw}} & \multicolumn{3}{c|}{\emph{filtered}} & \multirow{2}{*}{Avg.}
& \multicolumn{3}{c|}{\emph{raw}} & \multicolumn{3}{c|}{\emph{filtered}} & \multirow{2}{*}{Avg.} \\
Train $\downarrow$ & full & sent & mix   & full & sent & mix   &       & full & sent & mix         & full & sent & mix  \\
\midrule \midrule
full-\emph{raw} & 99.82 & 81.89 & 84.67 & 99.72 & 81.00 & 84.07 & 88.53
& 98.79 & 83.64 & 86.32 & 98.57 & 82.77 & 85.85 & 89.32\\
sent-\emph{raw} & 99.40 & 98.43 & 98.56 & 99.24 & 98.47 & 98.59 & \textbf{98.78}
& 97.76 & 95.75 & 96.11 & 97.68 & 95.31 & 95.77 & \textbf{96.40}\\
mix-\emph{raw} & 99.44 & 98.31 & 98.47 & 99.32 & 98.37 & 98.51  & 98.74
& 97.70 & 95.68 & 96.04 & 97.65 & 95.27 & 95.73 &  96.35\\
\midrule \midrule
full-\emph{filtered} & 99.82 & 87.17 & 89.05 & 99.79 & 86.60 & 88.67 & 91.85 & 98.25 & 91.04 & 92.30 & 98.14 & 91.15 & 92.48 & 93.89\\
sent-\emph{filtered} & 96.97 & 97.22 & 97.19 & 99.09 & 98.43 & 98.53 & \textbf{97.91} & 96.60 & 92.81 & 93.47 & 97.94 & 95.86 & 96.26 & 95.49 \\
mix-\emph{filtered} & 96.28 & 96.43 & 96.41 & 99.45 & 98.37 & 98.53 & 97.58
& 97.43 & 94.09 & 94.68 & 97.66 & 95.61 & 96.01 & \textbf{95.91}\\
\bottomrule
\end{tabular}
}
\vspace{0.8em}
\caption{F1 scores (\%) of RoBERTa models at full \& sent \& mix mode.}
\label{tab:result-full-sent-mix}
\end{table}

%% file: tables/qa_table.tex
\begin{table}[h!]
\centering
\setlength{\tabcolsep}{5pt}
\resizebox{0.98\textwidth}{!}{
\begin{tabular}{c|ccc|ccc|c|ccc|ccc|c}
\toprule[1.5pt]
& \multicolumn{7}{c|}{\bf English}   & \multicolumn{7}{c}{\bf Chinese}  \\
\midrule
\multirow{2}{*}{Test $\rightarrow$}
& \multicolumn{3}{c|}{\emph{raw}} & \multicolumn{3}{c|}{\emph{filtered}} & \multirow{2}{*}{Avg.}
& \multicolumn{3}{c|}{\emph{raw}} & \multicolumn{3}{c|}{\emph{filtered}} & \multirow{2}{*}{Avg.} \\
 & full & sent & mix   & full & sent & mix   &       & full & sent & mix         & full & sent & mix  \\
\midrule \midrule
 & \multicolumn{14}{c}{Train $\rightarrow$ \emph{raw} - full} \\ \midrule
\bf Single & 99.82 & 81.89 & 84.67 & 99.72 & 81.00 & 84.07 &  88.53    & 98.79 & 83.64 & 86.32 & 98.57 & 82.77 & 85.85 &  89.32   \\
\bf QA     & 99.84 & 92.68 & 93.70 & 99.75 & 92.34 & 93.46 &  95.30    & 98.99 & 80.56 & 83.85 & 98.73 & 80.24 & 83.89 &  87.71   \\
\midrule \midrule
 & \multicolumn{14}{c}{Train $\rightarrow$ \emph{filtered} - full} \\ \midrule
\bf Single & 99.82 & 87.17 & 89.05 & 99.79 & 86.60 & 88.67 &  91.85    & 98.25 & 91.04 & 92.30 & 98.14 & 91.15 & 92.48 &  93.89   \\
\bf QA     & 99.70 & 96.14 & 96.64 & 99.70 & 96.07 & 96.61 & \bf 97.48 & 97.29 & 92.10 & 93.01 & 97.18 & 92.40 & 93.31 & \bf 94.22 \\
\bottomrule
\end{tabular}
}
\vspace{0.8em}  
\caption{F1 scores (\%) of RoBERTa models trained with QA \& Single settings.}
\label{tab:result-qa}
\end{table}

%% file: tables/source_table.tex
\begin{table}[h!]
\centering
\setlength{\tabcolsep}{5pt}
\resizebox{0.98\textwidth}{!}{
\begin{tabular}{lc|cc|cc|cc|cc|cc}
\toprule[1.5pt]
Model & Test & F1-hu & F1-ch & F1-hu & F1-ch & F1-hu & F1-ch & F1-hu & F1-ch & F1-hu & F1-ch \\ \midrule
\multicolumn{12}{c}{\bf English} \\ \midrule \midrule
& & \multicolumn{2}{c}{finance}   & \multicolumn{2}{c}{medicine}  & \multicolumn{2}{c}{open\_qa} & \multicolumn{2}{c}{reddit\_eli5}    & \multicolumn{2}{c}{wiki\_csai} \\ \midrule
\multirow{2}{*}{\bf RoBERTa} & full & 99.34  & 99.28  & 99.69  & 99.62  & 99.53  & 98.60  & 100.00 & 100.00 & 96.59  & 96.37 \\
& sent & 78.84  & 85.84  & 84.06  & 80.45  & 70.74  & 26.78  & 77.27  & 93.31  & 68.91  & 84.12 \\
\multirow{2}{*}{\bf GLTR} & full & 97.50  & 97.37  & 98.28  & 97.96  & 92.68  & 82.20  & 98.22  & 99.40  & 95.76  & 95.72 \\
& sent & 46.60  & 75.26  & 45.41  & 61.72  & 42.01  & 17.81  & 38.12  & 87.05  & 39.24  & 76.94 \\
\midrule \midrule
\multicolumn{12}{c}{\bf Chinese} \\ \midrule
& & \multicolumn{2}{c}{finance}   & \multicolumn{2}{c}{law}  & \multicolumn{2}{c}{open\_qa} & \multicolumn{2}{c}{nlpcc\_dbqa}    & \multicolumn{2}{c}{baike} \\
\midrule
\multirow{2}{*}{\bf RoBERTa} & full & 98.87 & 97.99 & 97.78 & 98.50 & 98.75 & 99.33 & 97.42 & 95.42 & 94.61 & 93.99  \\
& sent & 95.00 & 80.46 & 93.77 & 86.23 & 91.17 & 93.77 & 90.10 & 63.29 & 86.08 & 88.88  \\
\multirow{2}{*}{\bf GLTR} & full & 86.67 & 80.42 & 82.41 & 88.89 & 85.75 & 93.15 & 77.25 & 69.78 & 81.62 & 77.91  \\
& sent & 36.91 & 32.80 & 33.99 & 46.22 & 36.45 & 75.21 & 46.39 & 27.50 & 48.10 & 71.72 \\
\bottomrule \\
\end{tabular}
}
\caption{Human (F1-hu) and ChatGPT (F1-ch) detection F1 scores (\%) w.r.t. different data source, models are trained on filtered full text, tested on filtered full and sent. On HC3-Chinese, we omitted the results of \emph{medicine} and \emph{psychology} domains, which are similar to \emph{finance} and \emph{open\_qa}, respectively.}
\label{tab:result-source}
\end{table}